\documentclass{article}

\usepackage{PRIMEarxiv}

\usepackage[natbibapa]{apacite}
\usepackage[utf8]{inputenc} 
\usepackage[T1]{fontenc}    
\usepackage{hyperref}       
\usepackage{url}            
\usepackage{booktabs}       
\usepackage{amsfonts}       
\usepackage{nicefrac}       
\usepackage{microtype}      
\usepackage{lipsum}
\usepackage{fancyhdr}       
\usepackage{graphicx}       
\graphicspath{{media/}}     
\usepackage{multirow}
\usepackage{tabularx}
\usepackage{amsmath}
\usepackage{microtype}

\makeatletter
\newcommand\footnoteref[1]{\protected@xdef\@thefnmark{\ref{#1}}\@footnotemark}
\newcommand\sbullet[1][.5]{\mathbin{\vcenter{\hbox{\scalebox{#1}{$\bullet$}}}}}
\makeatother

\title{Neural Data-to-Text Generation Based on Small Datasets: Comparing the Added Value of Two Semi-Supervised Learning Approaches on Top of a Large Language Model
}

\author{
  Chris van der Lee \\
  Tilburg University \\
  Tilburg center for Cognition \\ and Communication \\
  \texttt{c.vdrlee@uvt.nl} \\
   \And
  Thiago Castro Ferreira \\
  Universidade Federal de \\ Minas Gerais \\
  Faculdade de Letras \\
  \texttt{thiagocf05@ufmg.br} \\
  \And
  Chris Emmery \\
  Tilburg University \\
  Department of Cognitive Science \\ and Artificial Intelligence \\
  \texttt{c.d.emmery@uvt.nl} \\
  \AND
  Travis Wiltshire \\
  Tilburg University \\
  Department of Cognitive Science \\ and Artificial Intelligence \\
  \texttt{t.j.wiltshire@uvt.nl} \\
  \And
  Emiel Krahmer \\
  Tilburg University \\
  Tilburg center for Cognition \\ and Communication \\
  \texttt{e.j.krahmer@uvt.nl} \\
   \And
}

\begin{document}
\maketitle

\begin{abstract}
This study discusses the effect of semi-supervised learning in combination with pretrained language models for data-to-text generation. It is not known whether semi-supervised learning is still helpful when a large-scale language model is also supplemented. This study aims to answer this question by comparing a data-to-text system only supplemented with a language model, to two data-to-text systems that are additionally enriched by a data augmentation or a pseudo-labeling semi-supervised learning approach. Results show that semi-supervised learning results in higher scores on diversity metrics. In terms of output quality, extending the training set of a data-to-text system with a language model using the pseudo-labeling approach did increase text quality scores, but the data augmentation approach yielded similar scores to the system without training set extension. These results indicate that semi-supervised learning approaches can bolster output quality and diversity, even when a language model is also present.
\end{abstract}


\section{Introduction}

Neural NLG methods are notoriously data hungry, and rely on large-scale datasets which typically require large amounts of effort and resources to construct \citep{Gkatzia2016c}. The fact that such datasets are rare and difficult to develop creates a so-called \textit{data bottleneck} \citep{oraby_etal_2019}. Due to the lack of large datasets, many neural NLG approaches rely on relatively small datasets, which not only affects output quality, but also output diversity  \citep{holtzman_etal_2020}.

\begin{figure*}[t]
    \centering
    \includegraphics{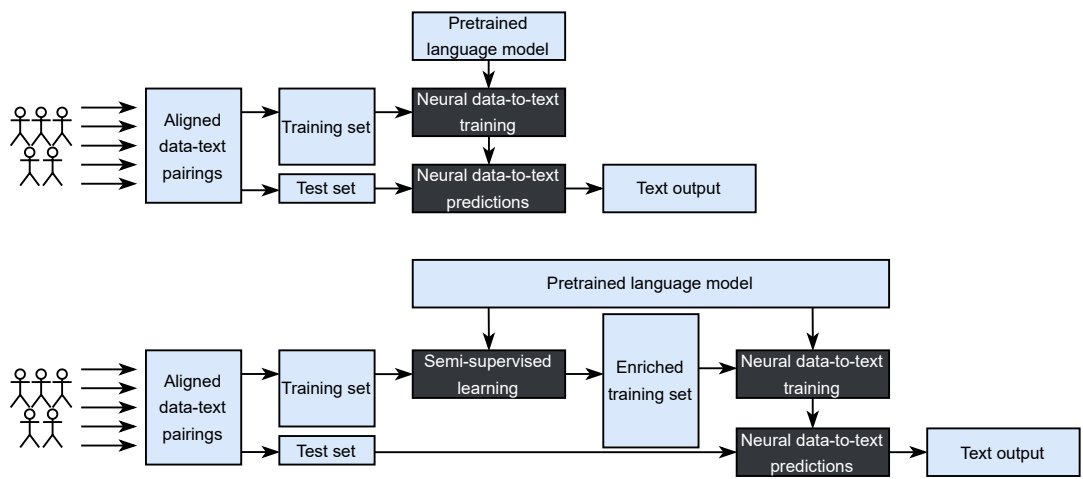}
    \caption{Schematic overview of supervised learning (top) vs. semi-supervised learning (bottom) for the current study.}
    \label{fig:schemessl}
\end{figure*}

One of the NLG subtasks that especially suffers from the consequences of small-scale datasets is data-to-text generation: the task of producing adequate, fluent and natural language text from non-linguistic structured data \citep{gatt_krahmer_2018}. (Supervised) neural data-to-text NLG involves the collection of parallel data-text datasets, aligning data and linguistic realisations of these data. However, collecting these datasets is difficult because sets of texts and corresponding data are not a common natural occurrence \citep{shimorina_etal_2019}. On the other hand, \textit{unpaired} texts and data are significantly more common and easily collected \citep{qader_etal_2019}. While these unpaired texts and data do not lend themselves for supervised data-to-text generation, they can be utilized by means of \textbf{semi-supervised learning}, the process of training a model on existing data-text pairings, and having this model create more synthetic pairings for the training set (see also \hyperref[fig:schemessl]{Figure~\ref*{fig:schemessl}}). 

Although more data generally leads to better performing machine learning systems, it is unclear to what extent a system benefits from including (possibly imperfect) synthetic data. Including imperfect synthetic data could result in cascading of errors, where an error early on impacts all later processes in the data-to-text conversion \citep{castro_ferreira_etal_2019}. However, previous data-to-text generation studies suggest that semi-supervised learning increases output quality compared to a supervised approach, especially when the labeled dataset size is small \citep[e.g.,][]{qader_etal_2019,schmitt_etal_2020,su_etal_2020,tseng_etal_2020,chang_etal_2021}. Besides improved output quality, adding more training examples using semi-supervised learning also might increase the language diversity of the output.

Recent developments in large-scale language models, utilizing the Transformers architecture \citep{vaswani_etal_2017} may offer an alternative solution to the data bottleneck. They leverage data from various domains to supplement the in-domain information that is available \citep{sun_etal_2020}. These language models have also been found to have a beneficial impact on output quality for the data-to-text task, just like semi-supervised learning \citep[e.g.,][]{kale_rastogi_2020}. Both of these approaches aim to improve the data-to-text training model by providing extra information in addition to the information that is present in the training set as an enriched training set. Therefore, it is conceivable that the massive amounts of information already incorporated in a language model makes the use of semi-supervised learning redundant. 

However, studies have also shown that the beneficial effect of a language model decreases when a model is overfit too much during finetuning \citep{greco_etal_2019}. Overfitting is also more likely to occur with small datasets. The fact that semi-supervised learning increases the dataset size may therefore help against overfitting. If so, language models and semi-supervised learning would be complementary approaches that might lead to better performance when used in conjunction. However, not much is known about the effect that this combination of language-models with semi-supervised learning has in a data-to-text setting. Furthermore, provided that the addition of semi-supervised learning is beneficial, the type of semi-supervision approach that leads to the best results is currently unknown, as experimental comparisons of this are scarce as well.

This study will investigate when and how semi-supervised learning affects the output diversity and quality when used in combination with language models for data-to-text generation. Two different semi-supervised learning approaches are investigated---both utilizing pretrained Transformers models \citep{vaswani_etal_2017}---and the impact that these approaches may have on the output quality and diversity of a neural data-to-text system with a language model. The semi-supervised learning approaches used in this study are: (I) a \textbf{data augmentation} approach, where several variants of a training text are generated by replacing certain words with synonyms or semantically similar words, (II) a \textbf{pseudo-label} approach, where unlabeled texts are given data labels by an information extraction (semantic parsing) model trained on the existing labeled training data. The synthetic data-text pairs obtained via these two approaches are then added to the original training set in a neural data-to-text system to generate new texts. 

Even though a handful of recent papers have started exploring the benefits of semi-supervised learning for data-to-text generation, this is the first study which presents a detailed, large scale analysis (1) of different corpora, in two different languages, (2) that systematically compares different methods for semi-supervised learning, also in combination with pretrained language models, and (3) that performs an exhaustive evaluation of the different methods, by combining automatic analyses with a human evaluation and an error analysis, in line with recent best practices for evaluation.

\subsection{Hypotheses}
Previous studies have consistently found that semi-supervised learning leads to improvements in output quality \citep[e.g.,][]{kulhanek_etal_2021,riabi_etal_2021,tandon_etal_2018,alberti_etal_2019, chang_etal_2021b,kedzie_mckeown_2019}. Additionally, \cite{sun_etal_2020} examined the dynamics between semi-supervised learning and language models in a text classification setting and found that a combination of the two approaches led to the highest classification scores. Therefore, we pose the following hypothesis:

\textbf{H1.} \textit{Extending the training set with semi-supervised learning increases the output \textbf{quality} of a neural data-to-text system with a language model (compared to a data-to-text system with a language model only trained on the base training set).}

Finetuning a text generation system with language models on small or non-diverse training data may lead to limited diversity in the output, even though language models themselves are trained on enormous amounts of text. This has to do with the known propensity for catastrophic forgetting that neural networks display \citep{greco_etal_2019}: the model is overfit too tightly during finetuning, which leads to ``forgetting'' about the capabilities of language modeling. Therefore, extending the finetuning dataset with more various language using semi-supervised learning could have a positive impact on diversity, which has also been found in a study by \cite{kulhanek_etal_2021}. Based on these results, we pose the following hypothesis:

\textbf{H2.} \textit{Extending the training set using semi-supervised learning increases the output \textbf{diversity} of a neural data-to-text system with a language model (compared to a data-to-text system with a language model only trained on the base training set).}

Although different semi-supervised learning approaches have shown their potential for various NLP/NLG tasks, not much is known about \textit{which} semi-supervised learning approaches are the most effective \citep{sun_etal_2020}. Therefore, we will pose the following exploratory research question:

\textbf{RQ1.} \textit{Do the data augmentation and pseudo-label approaches differ in terms of output quality and output diversity when used as semi-supervised learning approaches in a neural data-to-text system with a language model?}

Finally, we will investigate dataset and language as factors in this study. For datasets, we make a distinction between crowdsourced on the one hand \citep[e.g., WebNLG, E2E, ToTTo;][]{novikova_etal_2017b,gardent_etal_2017a,gardent_etal_2017b,parikh_etal_2020}, and datasets that are created from ``naturally occurring'' human-written texts on the other \citep[e.g., YelpNLG, RotoWire, CACAPO;][]{oraby_etal_2019,wiseman_etal_2017,vanderlee_etal_2020}. For the construction of a crowdsourced dataset, crowdsource workers write corresponding texts for a given set of meaning representations. This approach is reasonable for the construction of a large-scale dataset, provided that time and resources are not an issue. However, the very procedure of using crowdsource workers to verbalize a set of meaning representations without any given context, ensures that texts are mostly focused on high fidelity, while sacrificing on criteria like fluency and enjoyability \citep{vanderlee_etal_2020}. This means that output from systems trained on crowdsourced datasets is less likely to contain diverse language. Alternatively, it is possible to construct a dataset with texts written in real-world scenarios, rather than lab-setting verbalizations by crowdsource workers. This can be done by collecting texts and corresponding data on a large scale, without having detail-level alignment information \citep[as is done in][]{wiseman_etal_2017,wang_2019,puduppully_etal_2019}, or by manually extracting aligned data from texts \citep[as is done in][]{oraby_etal_2019,vanderlee_etal_2020}. While such an approach is likely to facilitate more diverse and fluent language, the difficulty of this alignment task makes large-scale dataset collection a daunting endeavor. 

In the current study we will compare the two most widely used data-to-text datasets: E2E \citep{novikova_etal_2017b} and WebNLG \citep{gardent_etal_2017a,gardent_etal_2017b}, that are both large-scale and crowdsource-based, and CACAPO \citep{vanderlee_etal_2020}, that is smaller-scale and based on real-world texts, in two languages (viz. English and Dutch). Previous studies have suggested that the benefits of semi-supervised learning are greater in more low resource scenarios \citep{chang_etal_2021}. Therefore, we pose the following hypothesis:

\textbf{H3.} \textit{The beneficial effect of semi-supervised learning for a neural data-to-text system with a language model on output \textbf{quality} is bigger for a small-scale dataset (CACAPO) than for a large-scale dataset (WebNLG and E2E).}

Focusing on diversity, datasets based on real-world texts generally contain more diverse language than crowdsourced datasets \citep{vanderlee_etal_2020}. This implies that crowdsourced datasets have more to gain from semi-supervised learning approaches' potential to introduce more language diversity to the training set of crowdsourced datasets. We pose the following hypothesis:

\textbf{H4.} \textit{The beneficial effect of semi-supervised learning for a neural data-to-text system with a language model on output \textbf{diversity} is bigger for a crowdsourced dataset (WebNLG and E2E) compared to a dataset based on real-word texts (CACAPO).}

It is a well-known fact that the majority of NLP developments are focused on English, with many other languages lagging behind in terms of support \citep{bender_etal_2021,riabi_etal_2021}. This is also the case for language models. While such language models exist for other languages, and multilingual variants of state-of-the-art models exist \citep[e.g., mBert, mT5;][]{devlin_etal_2019,xue_etal_2021}, the size of these models is generally much smaller \citep{bender_etal_2021}, and they are oftentimes missing functionality. For instance, mT5 was not pretrained on downstream tasks like T5 was \citep{xue_etal_2021}. This means that the benefits of language models in data-to-text generation is likely smaller for underrepresented languages, which in turn might mean that the beneficial effects of semi-supervised learning is greater, especially when we take into consideration that the size of non-English datasets is generally smaller as well \citep{riabi_etal_2021}. We therefore pose the following hypothesis:

\textbf{H5.} \textit{The beneficial effect of semi-supervised learning is greater when trained on a Dutch dataset (CACAPO Dutch) compared to an English dataset (WebNLG, E2E, and CACAPO English) for a neural data-to-text system with a pretrained language model.}\footnote{All these hypotheses have also been preregistered before conducting the study at \url{https://aspredicted.org/in665.pdf}, following the advice given by \cite{vanmiltenburg_etal_2021}.}

\section{Background}

\subsection{Semi-Supervised Learning}
The goal of semi-supervised learning is to train a model (partially) on synthetic data (as opposed to data that is created by humans), which may lead to a better trained machine learning model, and hence improved performance. \hyperref[fig:schemessl]{Figure~\ref*{fig:schemessl}} gives a schematic overview of semi-supervised learning and how it differs from supervised learning, which is the standard for neural NLG. This approach has steadily grown in popularity with the rise of data hungry neural models, and is considered especially useful when the (labeled) training set is small-scale. Within the NLG domain, we have seen applications of this approach in, for instance, Question Answering \cite[e.g.,][]{alberti_etal_2019,riabi_etal_2021}, and Text Simplification \citep[e.g.,][]{surya_etal_2019,zhao_etal_2020}. The semi-supervised approach has also gained traction in the context of data-to-text generation. This is mostly in the form of joint learning systems, where an NLG system (that converts meaning representations into text), and a Natural Language Understanding system (that converts text into meaning representations) are feeding each other more synthetic data in a loop. Such an architecture allows for both unlabeled texts as well as meaning representations without aligned text to be included into the training data. Some studies suggest that the use of a joint learning system led to improvements on various metrics compared to a supervised NLG system \citep[e.g.,][]{qader_etal_2019,schmitt_etal_2020,su_etal_2020,tseng_etal_2020,chang_etal_2021}.

The architecture of the current research differs from these previous studies as it utilizes unlabeled/unaligned data in a non-joint way. This approach is based on the assumption that it is easier to extract information from a text than to generate text that accurately represents information \citep{wiseman_etal_2017}. It should also be noted that previous data-to-text studies using semi-supervised approaches partition a segment of the dataset and detach the data from the texts to create unlabeled data. Or they use value swapping (i.e., pairing each key with a randomly sampled value collected from the set of all data samples to obtain new combinations of key-value pairs) to create extra unlabeled data. Our approach tries to more closely emulate the application of this task in a real-world setting by collecting and utilizing unaligned texts that are not found in the datasets. Furthermore, none of the previous studies used language models in the architecture of their data-to-text generation system and only one previous study utilized these language models for semi-supervised learning \citep{chang_etal_2021b}. 

\subsection{Language Models}
Transformers-based language models have particularly shown their viability for generation tasks that involve meaning manipulation (e.g., summarization, text simplification, and question answering), but studies also suggest that the inclusion of transformers-based language models can lead to improvements in output quality for the data-to-text generation task \citep{chen_etal_2020,kale_rastogi_2020,mager_etal_2020,ribeiro_etal_2021}. Furthermore, Transformers-based language models have been found to perform well on very small datasets, with examples existing of few-shot, one-shot or even zero-shot learning \citep{brown_etal_2020}. These impressive performances might suggest that including language models in the architecture of an NLG system might make semi-supervised learning approaches redundant. However, it should be noted that language models and most semi-supervised learning approaches utilize different types of data. While language models leverage data from an immense variety of different domains, semi-supervised learning approaches are generally focused on employing in-domain data \citep{sun_etal_2020}. Therefore, a combination of language models and semi-supervised learning approaches might enhance performance, rather than cancelling each other's improvements out. \cite{sun_etal_2020} finds support for this notion in a text classification context. They found that the largest performance gain was achieved when the two were combined \citep{sun_etal_2020}. The authors also compared different semi-supervised learning approaches (in-domain pretraining and pseudo-labeling) and found performance differences between the two. Similarly, the current study also compares the performance of two semi-supervised learning approaches, but in a data-to-text generation context: \textbf{pseudo-labeling} and \textbf{data augmentation}.

\subsection{Pseudo-Labeling}
One of the most common semi-supervised learning approaches is the pseudo-labeling approach, where unlabeled data is assigned labels by a model, thus forming a large labeled dataset that can be used to train a model. In the context of NLP, this task is equal to information extraction (also known as semantic parsing, or natural language understanding), where a meaning representation is parsed from a text. Most of the existing semi-supervised NLG systems have employed information extraction for the creation of synthetic training data \citep[e.g.,][]{qader_etal_2019,schmitt_etal_2020,su_etal_2020,tseng_etal_2020,chang_etal_2021}. However, most of these studies also apply pseudo-labeling without utilizing any language model. Besides its suitability for various generation tasks, Transformers-based language models have also shown their potential for information extraction. For example, various authors have applied BERT-based information extraction successfully on small datasets \citep[e.g.,][]{nguyen_etal_2019,zhangr_etal_2020}, and one of the best performers on the semantic parsing subtask of the WebNLG+ Challenge 2020 \citep{castro_ferreira_etal_2020_webnlg} was a parser that used T5 as a pretrained model \citep{agarwal_etal_2020}. Building on these previous findings, the current study also utilizes an approach to pseudo-labeling that includes a pretrained model.

\subsection{Data Augmentation}
Data augmentation refers to strategies that increase training examples without explicitly collecting new data \citep{feng_etal_2021}. This can be done either by adding slightly edited copies of existing data, or by creating new synthetic data from existing data \citep{sun_etal_2020}. There are various ways to perform data augmentation, most of which have also been successfully applied to a data-to-text generation context, such as (back) translation \citep{kulhanek_etal_2021,riabi_etal_2021}, text generation from modified meaning representations \citep[e.g.,][]{tandon_etal_2018,alberti_etal_2019, chang_etal_2021b}, and noise injection \citep{kedzie_mckeown_2019}. These approaches were all found to increase performance. A less common way of data augmentation in data-to-text generation is by using synonym replacement and text editing, which has effectively been applied for text classification \citep[e.g.][]{zhang_etal_2015}, and hate speech detection \citep{emmery_etal_2022,rizos_etal_2019}. However, recent advances in learning-based quality estimation metrics, such as RUSE \citep{shimanaka_etal_2018}, BertScore \citep{zhang_etal_2020}, MoverScore \citep{zhao_etal_2019}, and BLEURT \citep{sellam_etal_2020} try to gauge the semantic similarity of generated sentences compared to a gold standard using language models. The ability of these metrics to detect synonyms and semantically similar language does illustrate the viability of synonym replacement in (data-to-text) NLG as well, as instilled knowledge of semantically similar words and phrases is the most important part of data augmentation based on synonym replacement. In the current study, we will further investigate the potential for (the synonym/semantically similar replacement approach for) data augmentation using language models as a semi-supervised learning approach in data-to-text generation.

\section{Approach}

\begin{table}[t]
\centering
\footnotesize
\begin{tabular}{lrrr}
\toprule 
\bf CACAPO & \bf No. of & \bf No. of & \bf No. of \\
\bf Dutch & \bf instan- & \bf unique & \bf tokens \\
 & \bf ces & \bf MRs & \\
\midrule
No Extension & 7,367 & 6,590 & 110,391 \\
\midrule
Dat\_Aug (S) & 14,719 & 13,258 & 220,753 \\
Dat\_Aug (M) & 22,067 & 19,955 & 331,092 \\
Dat\_Aug (L) & 44,067 & 40,072 & 661,943 \\
Dat\_Aug (XL) & 80,537 & 73,272 & 1,212,192 \\
\midrule
Pseu\_Lab (S) & 13,626 & 11,727 & 201,894 \\
Pseu\_Lab (M) & 20,010 & 16,779 & 296,171 \\
Pseu\_Lab (L) & 32,465 & 25,778 & 479,852 \\
Pseu\_Lab (XL) & 57,251 & 42,286 & 844,537 \\
\bottomrule
\end{tabular}
\begin{tabular}{lrrr}
\toprule 
\bf CACAPO & \bf No. of & \bf No. of & \bf No. of \\
\bf English & \bf instan- & \bf unique & \bf tokens \\
 & \bf ces & \bf MRs & \\
\midrule
No Extension & 7,923 & 7,613 & 153,663 \\
\midrule
Dat\_Aug (S) & 15,822 & 14,962 & 308,655 \\
Dat\_Aug (M) & 23,718 & 22,306 & 463,641 \\
Dat\_Aug (L) & 47,391 & 44,230 & 928,203 \\
Dat\_Aug (XL) & 86,648 & 79,595 & 1,700,241 \\
\midrule
Pseu\_Lab (S) & 17,482 & 16,671 & 354,748 \\
Pseu\_Lab (M) & 27,221 & 25,010 & 555,519 \\
Pseu\_Lab (L) & 47,112 & 41,863 & 961,614 \\
Pseu\_Lab (XL) & 82,528 & 70,884 & 1,681,230 \\
\bottomrule
\end{tabular}
\begin{tabular}{lrrr}
\toprule 
\bf WebNLG & \bf No. of & \bf No. of & \bf \bf No. of \\
& \bf instan- & \bf unique & \bf tokens \\
 & \bf ces & \bf MRs & \\
\midrule
No Extension & 24,404 & 10,672 & 349,712 \\
\midrule
Dat\_Aug (S) & 48,732 & 32,086 & 690,545 \\
Dat\_Aug (M) & 73,059 & 52,225 & 1,031,404 \\
Dat\_Aug (L) & 146,018 & 112,625 & 2,053,905 \\
Dat\_Aug (XL) & 267,478 & 212,312 & 3,757,401 \\
\midrule
Pseu\_Lab (S) & 29,395 & 15,586 & 449,320 \\
Pseu\_Lab (M) & 34,310 & 20,336 & 548,731 \\
Pseu\_Lab (L) & 44,073 & 29,729 & 743,161 \\
Pseu\_Lab (XL) & 62,393 & 47,199 & 1,108,450 \\
\bottomrule
\end{tabular}
\begin{tabular}{lrrr}
\toprule 
\bf E2E & \bf No. of & \bf No. of & \bf No. of \\
 & \bf instan- & \bf unique &\bf tokens \\
 & \bf ces & \bf MRs & \\
\midrule
No Extension & 42,061 & 4,862 & 840,760 \\
\midrule
Dat\_Aug (S) & 83,853 & 43,555 & 1,683,644 \\
Dat\_Aug (M) & 125,644 & 81,708 & 2,526,518 \\
Dat\_Aug (L) & 251,017 & 196,285 & 5,054,910 \\
Dat\_Aug (XL) & 459,956 & 385,975 & 9,267,858 \\
\midrule
Pseu\_Lab (S) & 48,489 & 10,814 & 973,220 \\
Pseu\_Lab (M) & 54,917 & 16,065 & 1,105,611 \\
Pseu\_Lab (L) & 67,774 & 25,091 & 1,371,851 \\
Pseu\_Lab (XL) & 93,487 & 39,622 & 1,901,112 \\
\bottomrule
\end{tabular}
\caption{Size-related descriptives for the standard and semi-supervised training sets.}
\label{tab:e2edesc}
\end{table}

\subsection{Datasets}
For the data augmentation approach, it is beneficial to be able to locate the exact position where data was verbalized in the text, so that augmentations in the text that have to do with the data could easily be traced back and changed in the data as well. This ensures that augmented variants of texts also align with its data counterpart. Therefore, datasets were chosen for this experiment that included such enriched information. These are: (enriched) E2E \citep{castro_ferreira_etal_2021}, (enriched) WebNLG \citep{castro_ferreira_etal_2018}, and CACAPO \citep{vanderlee_etal_2020}. We used the original train/development/test splits for these corpora. More specific characteristics of the datasets are discussed in more detail below.

It should also be noted that we differentiate between domains for the CACAPO and WebNLG dataset (E2E is only one domain), as we believe that treating them separately will not only result in higher performance, but also provide more rich and detailed information about the performance of the various methods. Domains in CACAPO and WebNLG are inherently different due to imbalances that exist in the data (mostly in WebNLG) and due to the very nature of the reports. More specifically: the differences in richness of information they provide, and the complexity and diversity of the language that is used. To fully capture the effects that these domain differences have on the performances of different approaches, it is necessary to look at the sub-corpora.

Furthermore, we added synthetic data to the original dataset in various quantities. This was done to exploratively study the effects of semi-supervised learning in a more detailed fashion. It could for instance reveal a saturation point where adding more synthetic data stops improving performance, or that performance decreases when more synthetic data is added which indicates cascading of errors. Sizes and statistics are described in more detail below.

\subsubsection*{E2E \citep{novikova_etal_2017b}}
E2E is focused on the restaurant domain and contains English verbalizations of data, which were collected using crowdsourcing. The data for this dataset is stored in a key-value format, similar to CACAPO. The dataset is split in a training, development and test ratio of 76.5\%-8.5\%-15\%, respectively. Of the three datasets in this study, E2E is the largest in terms of sheer size: \hyperref[tab:e2edesc]{Table~\ref*{tab:e2edesc}} summarizes the basis statistics for E2E (and the other corpora used in this study). E2E contains 42,061 instances (that is: aligned data-text pairs), and 840,760 tokens. Furthermore, it contains 4,862 unique meaning representations (i.e. data elements), less than the other two corpora, which suggests that this dataset contains more repetition compared to WebNLG and CACAPO. 

As there is no information available on the origins of the data used for E2E, it is difficult to collect new data or find comparable restaurant descriptions online. Therefore, we used E2E+ \citep{roberti_etal_2020} as extra data for the pseudo-labeling approach. E2E+ is a modified version of E2E where all slot data is replaced with comparable data. For instance, \textit{food} data is replaced using the adjectival forms of countries and nations found on Wikipedia, and \textit{name} and \textit{near} are replaced with New York restaurant names found in the Entree dataset \citep{burke_etal_1997} (e.g. \textit{Blue Spice serves highly rated Chinese food.} becomes \textit{El Charro serves highly rated Timorese food}). To investigate the effect of the pseudo-labeling approach, four sizes of aligned data-text information were added to the original training set: small, medium, large, extra large. These contained 12.5\%, 25\%, 50\%, and 100\% of the E2E+ data, respectively, in case of the pseudo-labeling approach.

\subsubsection*{WebNLG \citep{gardent_etal_2017a,gardent_etal_2017b}}
WebNLG is collected in a similar crowdsourced manner as E2E and is derived from DBPedia properties. These properties are different from E2E and CACAPO data, as they are not stored in a key-value format, but as SVO-triples (subject-verb-object). Each of these properties is related to a particular category in DBPedia. For the enriched WebNLG dataset, these domains are: \texttt{Airport}, \texttt{Astronaut}, \texttt{Building}, \texttt{City}, \texttt{ComicsCharacter}, \texttt{Food}, \texttt{Monument}, \texttt{SportsTeam}, \texttt{University} and \texttt{WrittenWork}. While the dataset is smaller than E2E in terms of tokens and instances (24,404 instances; 349,712 tokens), it does seem more varied in its composition as evidenced by the number of unique meaning representations (10,672) (see \hyperref[tab:e2edesc]{Table~\ref*{tab:e2edesc}}). Furthermore, the dataset is split by a 60\%-20\%-20\% training, development, and test ratio. 

Following \cite{montella_etal_2020}, we collected Wikipedia texts as extra data for the pseudo-labeling approach for WebNLG. These texts are similar in nature, as Wikipedia pages are generally well-connected to the DBPedia variant of the page. Furthermore, Wikipedia texts are freely available and relatively easy to collect. For each of the DBPedia categories in the WebNLG dataset, we searched for similar overview pages on Wikipedia, and then scraped all the pages in the overview or in the subcategories of the overview.\footnote{The full list of pages that were collected can be found at \url{https://github.com/TallChris91/Extending\_Trainsets}.} Then, the summary (i.e., the first paragraph of the article) was taken from each page, split on a sentence-level, labeled, and added as extra data. The training set was extended with the sentences of 125, 250, 500, and 1000 summaries for respectively small, medium, large, and extra large. Since the pseudo-labeling data was split on a sentence-level, the original training set was also split on a sentence level to ensure consistency between the original training data and the input derived from the pseudo-labeling approach.

\subsubsection*{CACAPO \citep{vanderlee_etal_2020}}
The CACAPO dataset \citep{vanderlee_etal_2020} contains texts from the \texttt{Sports}, \texttt{Weather}, \texttt{Stocks}, and \texttt{Incidents} domain for both Dutch and English. Each domain contains information for 200 texts (1600 texts total), paired with manually annotated data for each sentence in a key-value format. It is split up in a 76.5\%-8.5\%-15\% training, development, and test ratio similar to E2E. Besides language differences, there are also topical differences between the English and Dutch part of the dataset: the \texttt{Weather} and \texttt{Stocks} report are relatively similar in their content, but the Dutch version of the \texttt{Sports} domain contains soccer reports, whereas the English version contains baseball reports \citep[based on][]{puduppully_etal_2019}. Similarly, the Dutch \texttt{Incidents} domain contains reports about traffic incidents \citep[from][]{Hendriks_2019}, while the English \texttt{Incidents} domain contains reports about firearm incidents \citep[see][for a detailed description]{vanderlee_etal_2020}. Size-wise both the Dutch and English datasets are the smallest in this study in terms of instances (Dutch: 7,367, English: 7,923) and tokens (Dutch: 110,391, English: 153,663) (see \hyperref[tab:e2edesc]{Table~\ref*{tab:e2edesc}}). The large number of meaning representations (Dutch: 6,590, English: 7,613) indicates a relatively large variation for its size.

Unlabeled texts for the pseudo-labeling approach were scraped using the same text collection methods as were used for the CACAPO dataset. This means that human-written texts were collected from the same selection of websites as were used for CACAPO. Furthermore, the texts were collected using an automatic scraper, or a tool that made saving texts in a correct format as effortless as possible, as was also done in the construction of CACAPO. Similar to WebNLG, the small version of the training set was extended with the sentences of 125 articles for small, 250 articles for medium, 500 articles for large, and 1000 for extra large.

\subsection{Data Augmentation}

\begin{table}[t]
\centering
\footnotesize
\begin{tabular*}{\textwidth}{l @{\extracolsep{\fill}} lrrrr}
\toprule
\textbf{Dataset} & \textbf{Domain} & \textbf{BLEU} & \textbf{BLEURT} & \textbf{BertScore} & \textbf{$\Delta$ Grammar} \\
\midrule
\multirow{8}{*}{CACAPO} & \texttt{Incidents} (EN) & 17.32 & 45.97 & 51.49 & +0.14 \\
 & \texttt{Sports} (EN) & 40.30 & 43.05 & 47.09 & +0.15 \\
 & \texttt{Stocks} (EN) & 22.83 & 48.41 & 46.92 & +0.14 \\
 & \texttt{Weather} (EN) & 30.21 & 47.55 & 45.67 & +0.10 \\
 & \texttt{Incidents} (NL) & 17.44 & 38.04 & 83.58 & +0.03 \\
 & \texttt{Sports} (NL) & 22.69 & 34.67 & 83.77 & +0.11 \\
 & \texttt{Stocks} (NL) & 20.11 & 35.75 & 84.18 & -0.06 \\
 & \texttt{Weather} (NL) & 26.78 & 43.63 & 83.11 & +0.19  \\
\midrule
\multirow{10}{*}{WebNLG} & \texttt{Airport} & 13.65 & 40.50 & 36.28 & +0.12  \\
 & \texttt{Astronaut} & 8.91 & 45.19 & 55.13 & +0.07  \\
 & \texttt{Building} & 11.39 & 43.63 & 42.15 & +0.12  \\
 & \texttt{City} & 9.26 & 44.23 & 40.69 & +0.04  \\
 & \texttt{ComicsCharacter} & 64.84 & 40.00 & 44.16 & +0.06  \\
 & \texttt{Food} & 41.11 & 44.06 & 43.90 & +0.13  \\
 & \texttt{Monument} & 16.26 & 42.86 & 47.63 & +0.07  \\
 & \texttt{SportsTeam} & 14.54 & 40.64 & 41.06 & +0.10  \\
 & \texttt{University} & 32.47 & 46.95 & 44.94 & +0.08  \\
 & \texttt{WrittenWork} & 20.86 & 42.84 & 40.39 & +0.10  \\
\midrule
E2E & & 24.27 & 46.78 & 57.99 & +0.09 \\
\bottomrule
\end{tabular*}
\caption{BLEU, BLEURT, BertScore, and mean difference in grammatical errors; comparing the original texts to the top 10 (XL-size) augmented sentences.}
\label{tab:dataugstats}
\end{table}



For data augmentation, we use lexical substitution \citep{mccarthy_etal_2007}; i.e., for specific words in the input (the target words) we determine multiple alternatives that are semantically similar (the substitution candidates). For this purpose, we use \cite{emmery_etal_2021}'s implementation of \cite{zhou_etal_2019}'s work. Under the framework of masked language modeling, to predict synonyms rather than any word fitting a particular (masked) position $t$, \cite{zhou_etal_2019} proposed using Dropout \citep{srivastava_etal_2014}. 

Given a document $D = (w_0, w_1, \ldots, w_t, \ldots, w_n)$, instead of masking position $t$, Dropout is applied to the BERT-internal embedding at position $t$. The intuition is that, rather than BERT predicting identical words when the original word's embedding is passed, the partly-zeroed embedding results produces synonyms instead. These are then the substitution candidates $C$, which we contextually re-rank using a similarity score:\footnote{In the original work, this is a subcomponent of the ranking function. We observed little difference in ranking by adding the word probability and $\alpha$ weightings (which do add computational complexity).}
\begin{equation}
 \textsc{sim}\left(D, D^{\prime} ; t\right) = \sum_{i}^{n} \alpha_{i, t} \times
 \Lambda\left(\boldsymbol{h}\left(D_{i}\right), \boldsymbol{h}\left(D_{i}^{\prime} \right) \right),
\end{equation}
where $\boldsymbol{h}\left(D_{i}\right)$ is the concatenation of BERT's last four layers for a given $i^{th}$ token in the original document $D$. The same concatenation is used for $\boldsymbol{h}\left(D'_{i}\right)$, but with target word $w_t$ from $D$  replaced with some candidate $c \in C$ at the index of $t$; i.e., $D' = (w_0, \ldots, c_t, \ldots w_n)$. Furthermore, $\Lambda$ denotes cosine similarity, and $\alpha_{i, t}$  the average self-attention score across all heads in all layers ranging from the $i^{th}$ token to the $t^{th}$ position in $D$. Finally, candidates are removed if they do not match certain criteria: their similarity scores should be > 0.9, and should not be punctuation or single characters, UNK tokens, plurals or capitalized versions of, or equal to the target word, subwords, or already exist in the sentence. BERT-large \citep{devlin_etal_2019} was used to generate the candidates for English, BERTje \citep{devries_etal_2019} for Dutch, and Dropout was set to 0.2.

For the target words, we chose all nouns, adjectives, adverbs, and numerals---tagged using SpaCy \citep{montani_honnibal_2017}. Similar to \cite{emmery_etal_2022}, we fill each position with a candidate simultaneously (i.e., using the highest ranked candidates for each target word to produce the first augmented instance, and so on; e.g., \textit{``What will the \textbf{weather} be like this \textbf{afternoon} in \textbf{Preston}?''} $\rightarrow$\space \textit{``What will the \textbf{air} be like this \textbf{evening} in \textbf{Manchester}?''}). We repeat this step for a maximum of twenty instances. If target words do not have up to the maximum amount of substitution candidates, they are left as the original words instead. The top 1 (small), 2 (medium), 5 (large), and 10 (extra large) instances of each text, based on their BERT similarity score with respect to the original sentence, were then added to the training datasets. As previously noted, enriched versions of corpora were used for data augmentation to ensure that augmentations were also applied to the aligned data.

We acknowledge the difficulty to measure the performance of data augmentation using automatic metrics, since most metrics are based on a comparison to a gold standard. Furthermore, language model-based semantic distance metrics (such as BertScore and BLEURT) are very similar in nature compared to the data augmentation approach used in this study, which might make their scores more akin to a manipulation check rather than an accurate reflection of semantic similarity between an augmented sentence and its original. Still, performing an evaluation using these metrics offers novel information as it measures sentence-level semantic consistency, which has not been measured in full during the data augmentation process. Therefore, we calculated the average BLEU \citep{papineni_etal_2002}, BLEURT \citep{sellam_etal_2020}, and BertScore \citep{zhang_etal_2020} scores of the augmented texts compared to their original. A lower BLEU score (a straightforward metric that measures text-overlap between a candidate and reference) and higher performance on BLEURT and BertScore (metrics that aim to measure semantic similarity between a candidate and reference) might suggest that texts have been augmented fundamentally, while still conveying a semantically similar message. Furthermore, we used LanguageTool\footnote{\url{https://languagetool.org}} to calculate the difference in grammatical errors compared to the original sentences, as an indicator for (relative) grammatical correctness. 

The results indicate that the data augmentation was generally effective. The BLEU scores are mostly around 10-20 for all domains, although a few exceptions exist. These low BLEU scores suggest that a large chunk of the original sentences were modified, meaning that the training data became more varied. We also see that the BLEURT and BertScore numbers are higher than BLEU for almost every domain, with scores in the 40-50 range for all domains. The BertScores for the Dutch domains are an exception, which rise above 80 due to the fact that rescaling with a baseline is not possible for this language. Nevertheless, these scores seem to indicate that the semantic similarity to the original text is kept relatively intact. Finally, the difference in grammatical errors is small, with generally only one to two tenths more errors being found in the augmented texts compared to the original texts. This suggests that data augmentation adds few new grammatical errors to the texts. Thus, overall, we can see these scores as an indicator that the data augmentation approach indeed manages to modify a sentence fundamentally, while still keeping the text relatively semantically similar to the original and relatively error free (see \hyperref[tab:dataugstats]{Table~\ref*{tab:dataugstats}}).

\begin{figure}[t]
\footnotesize

\rule[1ex]{\columnwidth}{0.1pt}

\centering
\begin{tabular}{ll}
\bf{Atribute} & \bf{Value} \\
Name & \textit{Wildwood} \\
eatType & \textit{pub} \\
food & \textit{Indian} \\
area & \textit{city centre} \\
familyFriendly & \textit{no} \\
near & \textit{Raja Indian Cuisine} \\
\end{tabular}
\\[3pt] \large{$\downarrow$}
\vspace{0.2em}
\\ \footnotesize{\textit{name @SEP@ Wildwood @EOF@ eatType @SEP@ pub @EOF@ food @SEP@ Indian @EOF@ area @SEP@ city centre @EOF@ familyFriendly @SEP@ no @EOF@ near @SEP@ Raja Indian Cuisine\newline}}
\rule[1ex]{\columnwidth}{0.1pt}
\caption{Example of attribute-value pairs and the corresponding data string. @SEP@ = seperator, @EOF@ = end of field.}
\label{tab:exampledata}
\end{figure}

\subsection{Pseudo-Labeling}

Similar to \cite{schmitt_etal_2020}, we framed the pseudo-labeling task as a text-to-text translation task as this approach could handle the differences in data formats between all three datasets most effortlessly and effectively (compared to, for instance, span labeling, or extractive question answering). While the most straightforward text-to-text translation purpose is to translate a text from, for instance, English to German, text-to-text translation can actually be used effectively for a multitude of Natural Language Processing tasks, as most of these involve conversion of one text format into another. T5 \citep{raffel_etal_2020} was developed with this purpose in mind. This architecture, also known as Text-to-Text Transfer Transformer, is a large pre-trained language model resulting from an empirical survey to determine which transfer learning techniques work best. Different from classification language models such as BERT, T5 works as a unified text-to-text-approach where all the NLP tasks are reframed such that its inputs and outputs are strings. This text-to-text framework is a multi-task one, sharing the parameters, loss function, and hyperparameters on any NLP task, including machine translation, document summarization, question answering, and classification tasks (e.g., sentiment analysis). To set the desired task for the model, a prefix needs to be inserted in the input such as ``translate English to German'' for the machine translation task or ``summarize'' for the summarization one.

In this case, we ``translated'' a ``data language'' to Dutch or English using T5-large \citep{raffel_etal_2020} for pretraining of the English pseudo-label model, and mT5-large \citep{xue_etal_2021} for pretraining of the Dutch pseudo-labeling model \citep[following][]{agarwal_etal_2020}. This was done using run\_translation.py from \url{https://github.com/huggingface/transformers} \citep{wolf_etal_2019} with 30 epochs and a batch size of 8. 
T5 and mT5 were further finetuned on the original training- and development set of the CACAPO, E2E, and WebNLG datasets and applied to the test set to calculate performance. Furthermore, the trained models were applied to the unlabeled new texts that were collected to extend the training set. 

More specifically, we converted the data into a structured string format that follows the data structure of the dataset (see \hyperref[tab:exampledata]{Figure~\ref*{tab:exampledata}}) using ``translate Dutch [resp. English] to Data: [...]'' as the prefix command. For the output, this string format was then converted back into structured data using a simple rule-based script. 

\begin{table}[t]
\centering
\footnotesize
\begin{tabular*}{\textwidth}{l @{\extracolsep{\fill}} lrrr|rrr}
\toprule 
 &                  & \multicolumn{3}{c|}{\bf Dev} & \multicolumn{3}{c}{\bf Test}\\
 & & \multicolumn{1}{l}{\textbf{P}} & \multicolumn{1}{l}{\textbf{R}} & \multicolumn{1}{l|}{\textbf{F1}} & \multicolumn{1}{l}{\textbf{P}} & \multicolumn{1}{l}{\textbf{R}} & \multicolumn{1}{l}{\textbf{F1}}\\
\midrule
\multirow{8}{*}{CACAPO} & \texttt{Incidents} (NL)   & 83.00     & 85.17  & 84.07 & 74.27     & 78.76  & 76.45 \\
 & \texttt{Sports} (NL)      & 73.65     & 77.97  & 75.75 & 74.33     & 76.03  & 75.17 \\
 & \texttt{Stocks} (NL)      & 85.96     & 89.17  & 87.54 & 90.37     & 89.72  & 90.04 \\
 & \texttt{Weather} (NL)     & 81.16     & 87.63  & 84.27 & 85.57     & 89.92  & 87.69 \\
 & \texttt{Incidents} (EN)   & 79.63     & 82.43  & 81.01 & 77.85     & 79.84  & 78.83 \\
 & \texttt{Sports} (EN)      & 79.02     & 80.07  & 79.54 & 79.95     & 79.95  & 79.95 \\
 & \texttt{Stocks} (EN)      & 80.60     & 84.84  & 82.66 & 83.08     & 79.27  & 81.13 \\
 & \texttt{Weather} (EN)     & 83.61     & 80.94  & 82.25 & 79.83     & 82.77  & 81.27 \\
\midrule
\multirow{10}{*}{WebNLG} & \texttt{Airport}          & 89.70     & 88.76  & 89.23 & 91.03     & 89.85  & 90.43 \\
 & \texttt{Astronaut}        & 96.13     & 95.41  & 95.77 & 97.00     & 94.95  & 95.97 \\
 & \texttt{Building}         & 89.81     & 91.04  & 90.42 & 89.86     & 89.31  & 89.59 \\
 & \texttt{City}             & 73.61     & 73.61  & 73.61 & 63.02     & 28.63  & 39.37 \\
 & \texttt{ComicsCharacter}  & 95.96     & 98.17  & 97.05 & 96.53     & 95.59  & 96.06 \\
 & \texttt{Food}             & 87.67     & 88.56  & 88.11 & 89.02     & 88.39  & 88.70 \\
 & \texttt{Monument}         & 72.38     & 68.88  & 70.59 & 52.83     & 50.76  & 51.77 \\
 & \texttt{SportsTeam}       & 81.52     & 81.79  & 81.65 & 88.07     & 88.22  & 88.15 \\
 & \texttt{University}       & 95.52     & 93.43  & 94.46 & 93.20     & 91.13  & 92.15 \\
 & \texttt{WrittenWork}      & 93.35     & 93.23  & 93.29 & 92.89     & 91.48  & 92.18 \\
\midrule
E2E &               & 85.65     & 91.74  & 88.59 & 88.17     & 82.72  & 85.36 \\
 \bottomrule
\end{tabular*}
\caption{Precision, Recall, and F1 scores of information extracted by our pseudo-labeling system.}
\label{tab:prf1scores}
\end{table}

We evaluated the performance of the pseudo-labeling approach by calculating the precision, recall, and micro-averaged F1 score on the development and test sets of all datasets. While we believe that these measures give a robust indication of the labeling quality, it should be noted that the model might not generalize well to the unlabeled texts, especially when the unlabeled texts are highly dissimilar from the texts seen in training (for instance, the pseudo-labeling model trained on E2E showed a considerable drop-off on the E2E+ data\footnote{On the synthetic data, P 54.07, R 60.93, and F1: 57.30 was achieved.}). Overall, the scores indicate that this pseudo-labeling approach performs well, with F1 scores well above the 70s for CACAPO and even in the high 80s and 90s for WebNLG and E2E (see \hyperref[tab:prf1scores]{Table~\ref*{tab:prf1scores}}). Two notable exceptions are the \texttt{City} and \texttt{Monument} domains for WebNLG that achieve much lower scores than other domains. This is likely caused by imbalanced data in the WebNLG dataset, which is especially prevalent in the \texttt{City} and \texttt{Monument} domain.

\subsection{Data-to-Text Generation}

The data-to-text approach utilized in this study was a neural end-to-end architecture where a set of input data is directly converted into English or Dutch text. This was done using Any2Some\footnote{\url{https://github.com/ThiagoCF05/Any2Some}}, which uses language models from the HuggingFace API \citep{wolf_etal_2019} to perform data-to-text generation, while offering advantages such as automatically clustering verbalizations based on the same data. Similar to the pseudo-labeling step, we used T5-large \citep{raffel_etal_2020} for the data-to-text conversion step as well. This time, using the text-to-text nature of the language model to perform data-to-text generation using ``Verbalize: [...]'' as the prefix command. As mentioned previously, T5 has been developed as an approach capable of handling a multitude of Natural Language Processing tasks where the inputs and outputs are reframed as strings. For this system, the input and output used in the pseudo-labeling step was essentially reversed: the data was again converted to a structured string format but this time used as input, with English or Dutch text serving as output. Previous research has suggested that T5 is a capable language model for the data-to-text generation task \citep{kale_rastogi_2020}. 

The model was finetuned on all individual domains for 16 epochs with a learning rate of 1e-5, early stopping of patience 5 and batch size of 2. Input and output strings were trimmed to a maximum size of 180 sub-tokens. For the Dutch CACAPO domains we used mT5-large, with the same hyperparameters except for 50 epochs. More epochs were necessary for this model to be properly trained, as mT5 was not trained on downstream tasks. Some examples of the input and output of the data-to-text generation system for each dataset and semi-supervised learning method can be found in \hyperref[tab:exampleoutputs]{Table~\ref*{tab:exampleoutputs}}.

\begin{table}[t]
\centering
\scriptsize
\begin{tabularx}{\textwidth}{lX}
\toprule
\textbf{BLEU} & Measures exact word match precision between model output and one or more references. \\[5pt]
\textbf{NIST} \citep{doddington_2002} & Similar to (corpus) BLEU, but adds more weight to more rare words. \\[5pt]
\textbf{METEOR} \citep{banerjee_lavie_2005} & Measures precision and recall of exact word matches between a reference and a candidate, also adds stemming and synonym matching. \\[15pt]
\textbf{ROUGE-L} \citep{lin_2004} & Looks at the Longest Common Subsequence between model output and one or more references and calculates the F1 score. \\[5pt]
\textbf{BertScore} & Measures the F1 score or the similarity between model output and one or more references, instead of exact matches, it computes similarity using contextual embeddings. \\
\bottomrule
\end{tabularx}
\caption{Definitions of the automatic metrics for text quality used in this study.}
\label{listofautmetrics}
\end{table}

\section{Evaluation}

The goal of this evaluation study was to investigate the contribution of the semi-supervised learning methods in data-to-text NLG. To investigate this, the evaluation study consisted of three parts: an automatic evaluation, a quantitative human evaluation, and an error analysis.\footnote{Ethical clearance was obtained from the \href{mailto:tshd.redc@tilburguniversity.edu}{Tilburg \texttt{University} School of Humanities and Digital Sciences Research Ethics and Data Management Committee} for this experiment (code: 2019.40). Furthermore, the study was pregistered at \url{https://aspredicted.org/in665.pdf} and the results of this study are available via \url{https://figshare.com/s/3959076f2d69d1381ccc}.} We aimed to follow the best practice guidelines as described in \citep{vanderlee_etal_2021} as much as possible in the setup and reporting of the evaluation study. 

For the automatic evaluation, multiple metrics were used to estimate output quality and output diversity. The quantitative human evaluation experiment measured aspects of text quality to further determine the performance of the different semi-supervised approaches relative to each other, and finally an error analysis was performed on the 15 worst scoring sentences per \textit{dataset} $\times$ \textit{semi-supervised learning approach} combination to investigate the shortcomings and challenges for each semi-supervised learning approach.

\subsection{Automatic Evaluation}
The performance of the three types of semi-supervised learning that were investigated in this research (no extension, data augmentation, and pseudo-labeling) was first tested for all domains in the CACAPO, E2E, and WebNLG dataset using automatic metrics that measure text quality and diversity. The text quality metrics served as a first test for H1, H3, H5, and RQ1. The text quality metrics employed in this study are displayed in \hyperref[listofautmetrics]{Table~\ref*{listofautmetrics}}.


Furthermore, we employ the diversity metrics based on \cite{vanmiltenburg_etal_2018}. These metrics are used to test H2 and H4, as they provide an objective and complete image of the diversity in the output of the systems, which cannot be measured as accurately with sentence/phrase-level human evaluation. The diversity metrics used in this study are described in \hyperref[listofdivmetrics]{Table~\ref*{listofdivmetrics}}.

\begin{table}[t]
\centering
\scriptsize
\begin{tabularx}{\textwidth}{lX}
\toprule
\textbf{Average sentence length} (ASL) & Average number of tokens per sentence. \\[5pt]
\textbf{Standard deviation of the sentence length} (SDSL) & How much variation there is in the number of tokens per sentence. \\[5pt]
\textbf{Number of types} (Types) & Number of unique word types in the output. \\[5pt]
\textbf{Mean segmented type-token ratio} (TTR 1) & Divides the generated texts into equal segments of a given token length (here: 100 tokens) and calculates the average type-token ratio of all these segments. \\[15pt]
\textbf{Bigram TTR} (TTR2) & Average type-token ratio of bigram types per 100 bigram tokens. \\[5pt]
\textbf{Percentage of novel texts} (\%Novel) & Texts generated by the system that do not occur in the training and development data. \\[5pt]
\textbf{Coverage} (Cov) & The percentage of learnable words (i.e., words in the original training or development set) that are recalled in the generated output. \\[15pt]
\textbf{Novelty} (Nov) & The percentage of novel words (i.e., words that do not appear in the original training or development set) that are in the generated output. \\[15pt]
\textbf{Local Recall} (Loc1) & The percentage of important words (i.e., adjectives, verbs, nouns, and adverbs) in a given test set text, that are recalled by the system's generated text. \\
\bottomrule
\end{tabularx}
\caption{Definitions of the automatic metrics for text diversity used in this study.}
\label{listofdivmetrics}
\end{table}


\subsection{Quantitative Human Evaluation}
\subsubsection*{Participants}
Participants of this study were recruited via Prolific, a crowdsourcing platform. For participation, participants received \$4.80 (the recommended amount according to the platform). In total, 193 people participated in the study, which was divided up in a Dutch version and an English version. In the recruitment phase, only participants were recruited that were native Dutch located in the Netherlands for the Dutch version, and native English speakers located in the United States for the English version. This resulted in 41 participants in the Dutch version, of which the majority were men (56\%) between the ages of 18 and 34 (90\%). Furthermore, the majority of the Dutch sample had a university (of applied sciences) degree (79\%). For the English version, 152 people participated in the study. The majority were women (64\%), roughly half of the participants were between the ages of 18 and 34 (48\%), and the majority have attended or completed college (87\%).

\subsubsection*{Design}
To ensure we captured the variety found in all datasets and among all semi-supervised learning approaches, we measured the text quality of outputs from the 3 investigated semi-supervised learning approaches on all 19 domains in the datasets we used (CACAPO: 8, E2E: 1, WebNLG: 10). We randomly sampled a total of 40 items per semi-supervised learning approach-domain combination, leading to a total of $19 \times 3 \times 40 = 2,280$ trials. Each trial was judged a total of 5 times to obtain a stable judgement of the trial.

Each participant was randomly assigned to a dataset domain. Dutch-speaking participants were randomly assigned to 1 of 4 domains (the four Dutch CACAPO domains), while English-speaking participants were randomly assigned to 1 of the other 15 (English) domains. Furthermore, each dataset domain had 2 versions, with each version containing 60 outputs total from the systems trained on the XL data (20 per semi-supervised learning approach) that were all not present in the other version. This amount of outputs was chosen to ensure that the sample contained enough variety to be representative of the variation in the full dataset, while the number of stimuli presented to individual participants was still manageable for them. Participants were randomly assigned to 1 of the 2 versions. The 60 outputs were presented in random order to compensate for potential order- or fatigue effects.


\subsubsection*{Procedure}
A survey was created using the \textit{Qualtrics} platform. First, a general introduction of the experiment and a consent form was given to the participants. After consenting to participate in the research, the participants were given detailed instructions about the experiment they were about to participate in. These instructions included guidelines on how to read the data input and the output texts, and how to rate said output texts. Furthermore, definitions were given for the scales they had to rate, and examples were given about good output texts and bad output texts. Instructions, guidelines, definitions, examples, as well as the questions themselves (as shown below), were translated to Dutch for the Dutch version of the evaluation to ensure that monolingual Dutch participants were able to comprehend the contents.

After these instructions, participants were asked to provide some demographical information and then the experiment started. Participants were shown a table with the original input data accompanied by a generated text from the NLG system trained on data-text pairings from one of the three semi-supervised learning approaches (no extension, data augmentation, pseudo-labeling). A selection was made of inputs that contained between 2-6 data elements, to keep the input data relatively understandable for participants. The generated texts were one sentence long (for CACAPO and WebNLG), or a few sentences long (between 1 and 6; for E2E). After viewing the input data and output texts, participants were asked to rate the texts on multiple items. Definitions of each item could also be found by hovering over the item. 

We measured \textit{fluency} using four seven-point Likert scale items based on \cite{sundar_1999,clerwall_2014} (consistency was high, with $\alpha$ = .97). The items were introduced by ``This sentence/short text is...'', followed by ``Clear'' (The overall message of the sentence/short text is clear.), ``Coherent'' (It is easy to follow the connections in the sentence/short text. The different pieces of information are connected in a correct way.), ``Understandable'' (The sentence/short text is written in a way that is easy to understand. There are no strange word choices or phrases that make the sentence/short text confusing.), and ``Well-written'' (The sentence/short text is fluent and easy to read.).

\textit{Correctness} was measured using three seven-point Likert scale items based on \cite{hoorn_vanwijngaarden_2010} (consistency was high, with $\alpha$ = .92). The items were introduced by ``Based on the data table, the information in this sentence/short text is...'', followed by ``Factual'' (The sentence/short text only describes the data in the data table. There is no extra information being described in the sentence/short text that is not represented in the data table.), ``Accurate'' (The information in the data table is represented correctly in the sentence/short text. There are no mistakes in the names and numbers, for instance.), ``Complete (All the (important) information from the data table is represented in the sentence/short text. There is no information missing in the sentence/short text that is represented in the data table.)''.

\begin{table}[t]
\centering
\scriptsize
\begin{tabularx}{\textwidth}{X}
\toprule
$\sbullet[.75]$ Does the text contain information that is not reported in the data table? \\
$\sbullet[.75]$ Is the text missing information that is in the data table? \\
$\sbullet[.75]$ Did you find any mistake involving the references? (e.g., ``Indian cuisine'' in the data table becomes ``German cuisine'' in the text, or the reference is not explicitly mentioned: ``He is the leader of the country.'' instead of ``Joe Biden is the leader of the United States of America.'') \\
$\sbullet[.75]$ Did you find any mistake involving the verb form? (e.g., The boy ``play'' soccer instead of ``plays'', or plainly missing a verb: ``The boy soccer'') \\
$\sbullet[.75]$ Did you find any mistake involving the determiners? (e.g., ``An'' boy.) \\
$\sbullet[.75]$ Did you find any mistake involving the punctuation or capitalization in the sentence? (e.g., ``The,boy is here;'' or ``the Boy Is here'') \\
$\sbullet[.75]$ Did you find any mistake involving strange lexical choices? (e.g., The player ``shot'' the goal.) \\
$\sbullet[.75]$ Did you find any mistake involving illogical/unnecessary repetition of words or phrases? (e.g., ``The the the'' boy) \\
$\sbullet[.75]$ Did you find any mistake involving connections between data points? (e.g., ``The leader of the cheeseburger is Barack Obama.'') \\
$\sbullet[.75]$ Is the sentence (or: one of the sentences) missing important parts to make it a full sentence? (e.g., ``The maximum temperature is.'' instead of ``The maximum temperature is 12 degrees Celsius.'') \\
Other (specify below) \\
\bottomrule
\end{tabularx}
\caption{Questions asked in the error analysis study.}
\label{erroranalysisitems}
\end{table}

\textit{Grammaticality} was measured using one multiple-choice question containing 4 options, based on \cite{ross_1979}. The question was introduced by: ``How grammatically correct is this sentence/short text?'' followed by (1) ``The sentence/short text sounds perfect. I would use it without hesitation.'', (2) ``The sentence/short text is less than perfect – something in it just doesn’t feel comfortable. Maybe lots of people could say it, but I never feel quite comfortable with it.'', (3) ``Worse than 2, but not completely impossible. Maybe somebody might use the sentence/short text, but certainly not me. The sentence/short text is almost beyond hope.'', (4) ``The sentence/short text is absolutely out. Impossible to understand, nobody would say it. Un-English.'' For the results section, this domain was reverse-coded to make the scores better interpretable. 

Finally, after rating all sentences/short texts, participants were fully briefed on the goal of the study, reminded of the contact addresses if they had more questions about the research, and thanked again for participation.

\subsection{Error Analysis}\label{erroranalysismethod}
An error analysis was performed to get a better understanding of the exact errors that can be found in the NLG output, which in turn may help to improve the various systems. The 15 worst-scoring texts (on the average of all three measured constructs) in the quantitative human evaluation experiment for each system-dataset combination (a total of 180 texts) were analyzed by 7 human annotators, all experts in language and communication and proficient in both Dutch and English; none having previously seen the output of the various systems. The annotators all coded 19 sentences jointly and 23 sentences each individually. Cohen's Kappa for multiple raters \citep{davies_fleiss_1982} was calculated for the jointly annotated part,\footnote{Using NLTK's \texttt{multi\_kappa} function} resulting in $\kappa$ = .45. This indicates moderate agreement \citep{landis_koch_1977}. 11 error categories for these 180 texts were developed based on \citep{castro_ferreira_etal_2019}, and less straightforward error categories were accompanied by a short description with examples. See \hyperref[erroranalysisitems]{Table~\ref*{erroranalysisitems}} for an overview of the items. If annotators selected ``Other'', they were able to input text to describe the error category they felt they encountered.



\begin{table}[t]
\centering
\footnotesize
\begin{tabular*}{\textwidth}{l @{\extracolsep{\fill}} lrrrrr}
\toprule
\bf Dataset     & \bf Train type & \bf BLEU  & \bf NIST & \bf BertScore & \bf METEOR & \bf ROUGE-L \\
\midrule
\multirow{3}{*}{CACAPO (en)} & No\_Ext       & 30.50  & 6.77 & 59.51     & 56.05  & 51.34   \\
 & Dat\_Aug    & 24.37 & 6.30  & 52.15     & 48.80   & 45.89   \\
 & Pseu\_Lab    & \textbf{36.21} & \textbf{7.55} & \textbf{63.83}     & \textbf{59.93}  & \textbf{56.55}   \\
\midrule
\multirow{3}{*}{CACAPO (nl)} & No\_Ext       & 33.94 & 6.77 & 84.60      & 52.91  & 51.97   \\
 & Dat\_Aug    & 38.30  & 7.56 & 86.86     & 59.01  & 58.31   \\
 & Pseu\_Lab    & \textbf{54.25} & \textbf{9.30}  & \textbf{89.84}     & \textbf{68.74}  & \textbf{68.05}   \\
\midrule
\multirow{3}{*}{E2E}         & No\_Ext       & \textbf{66.05} & \textbf{7.08} & \textbf{79.4}      & \textbf{80.21}  & \textbf{44.97}   \\
 & Dat\_Aug    & 28.41 & 4.15 & 56.41     & 62.49  & 33.40    \\
 & Pseu\_Lab    & 50.51 & 4.65 & 63.12     & 60.39  & 38.48   \\
\midrule
\multirow{3}{*}{WebNLG} & No\_Ext       & \textbf{47.91} & \textbf{8.74} & \textbf{71.3}      & \textbf{71.57}  & \textbf{59.88}   \\
 & Dat\_Aug    & 27.71 & 5.95 & 52.75     & 53.23  & 45.33   \\
 & Pseu\_Lab    & 44.55 & 8.32 & 67.82     & 68.74  & 56.82  \\
\bottomrule
\end{tabular*}
\caption{Automatic metric results of the different (XL-format) semi-supervised learning approaches (No\_Ext = no training set extension, Dat\_Aug = data augmentation, Pseu\_Lab = pseudo-labeling) for each dataset (bold = highest).}
\label{tab:autmetsperdataset}
\end{table}

\section{Results}
\subsection{Automatic Evaluation}
\subsubsection*{Automated Metrics for Text Quality}

The automatic analysis results are summarized in \hyperref[tab:autmetsperdataset]{Table~\ref*{tab:autmetsperdataset}} for the overall datasets, and presented per domain in \hyperref[tab:autmetsslm]{Table~\ref*{tab:autmetsslm}}. Additionally, \hyperref[BLEU_Fig]{Figure~\ref*{BLEU_Fig}} shows the effects of increasing the synthetic data from the semi-supervised learning methods on BLEU scores. Inspection of \hyperref[tab:autmetsperdataset]{Table~\ref*{tab:autmetsperdataset}} reveals a clear pattern: for CACAPO the pseudo-labeling approach consistently leads to the highest automatic metric scores, while for E2E and WebNLG data extension does not lead to better automatic scores overall. This pattern is consistent amongst all metrics, but we will zoom in on the BLEU differences in this section.

The results of the automatic metrics suggest that clear differences between datasets exist in the output quality achieved with the different semi-supervised learning approaches. For the English CACAPO dataset, BLEU scores, for instance, improved by 5.71 on average with the pseudo-labeling approach (30.50 to 36.21), but the data augmentation approach led to an average BLEU decrease of 6.13 compared to no extension (30.50 to 24.37). The positive effect on automatic metric scores was more noticeable for the Dutch part of the CACAPO dataset, where BLEU scores increased with 20.31 on average for the pseudo-labeling approach (33.94 to 54.25) and a 4.36 BLEU improvement for the data augmentation approach compared to the no extension approach (33.94 to 38.30). The most extreme improvement was reached for the Dutch weather dataset (a 52.38 improvement for pseudo-labeling compared to no extension; see \hyperref[tab:autmetsslm]{Table~\ref*{tab:autmetsslm}}). It is possible that the small size and limited vocabulary of the original weather training set was insufficient for a neural NLG system to be properly trained on, whereas it was when the extended training set from the semi-supervised learning approaches were applied. \hyperref[BLEU_Fig]{Figure~\ref*{BLEU_Fig}} also shows that increasing the amount of data leads to higher BLEU scores for both Dutch and English CACAPO; that is, for the pseudo-labeling method. This provides further support for the notion that extending the training set has a positive impact on output quality.

The pseudo-labeling approach for WebNLG generally led to a decrease in automatic metric scores (compared to no extension) albeit relatively small, with an average BLEU decrease of 3.36. However, the BLEU decrease for the data augmentation approach (compared to no extension) was more noticeable, with a 20.20 BLEU decrease. Decreases in automatic metric scores for the semi-supervised learning approaches compared to no extension were also observed for E2E: a 15.54 BLEU score difference for pseudo-labeling and a 37,64 decrease for data augmentation (see \hyperref[tab:autmetsperdataset]{Table~\ref*{tab:autmetsperdataset}}). Interestingly, the pseudo-labeling approach performed well for the \texttt{City} and \texttt{Monument} domains where F1 scores in the pseudo-labeling step were noticeably worse (see \hyperref[tab:autmetsslm]{Table~\ref*{tab:autmetsslm}}). This supports the notion that semi-supervised learning is mostly effective in situations where the original training set is small (as stated in H3), filling in the training data deficit by providing more examples. The general tendency, however, is that an increase in synthetic training data leads to a decrease in BLEU scores for both E2E and WebNLG, and for both pseudo-labeling as data augmentation (as shown in \hyperref[BLEU_Fig]{Figure~\ref*{BLEU_Fig}}). This decrease could suggest a cascading of errors where issues in the synthetic data negatively impact the quality of the generated output. Alternatively, it could be that the texts in E2E and most WebNLG domains are relatively homogeneous. Introducing more deviations from these texts increases the diversity of the output, which results in lower scores on automatic metrics. This could also explain the relatively low scores of the data augmentation approach, where the quantity of the deviations may have the biggest impact on the heterogeneity of the output. This possibility is further examined with the diversity metric scores, the quantitative human evaluation, and the error analysis. 

In any case, while these automatic metrics do not provide all-around support for the notion that semi-supervised learning combined with a language model increases output quality compared to an NLG system with a language model that is only finetuned on the original dataset (H1), the results are in line with H3: the beneficial effect of semi-supervised learning in the text quality metrics is only noticeable for the dataset (categories) that are small-scale or unbalanced. They are also in line with H5: the highest increase of the semi-supervised learning methods was observed for the Dutch CACAPO dataset.

\subsubsection*{Automated Metrics for Text Diversity}

\begin{table}[t]
\centering
\footnotesize
\setlength{\tabcolsep}{3pt}
\begin{tabular*}{\textwidth}{l @{\extracolsep{\fill}} lrrrrrrrrr}
\toprule
\bf Dataset     & \bf Train type & \bf ASL   & \bf SDSL & \bf Types & \bf TTR1 & \bf TTR2 & \bf \%Novel & \bf Cov  & \bf Nov  & \bf Loc1 \\
\midrule
\multirow{3}{*}{CACAPO (en)} & No\_Ext       & 17.26 & 8.29 & 3502  & 0.66 & 0.93 & 98.24  & 0.58 & 0.20 & 0.53 \\
 & Dat\_Aug    & 17.39 & 8.54 & \textbf{3797}  & \textbf{0.68} & \textbf{0.95} & \textbf{99.80}  & \textbf{0.61} & \textbf{0.24} & 0.51 \\
 & Pseu\_Lab    & \textbf{17.56} & \textbf{9.21} & 3709  & 0.67 & 0.93 & 98.50  & 0.60 & 0.22 & \textbf{0.57} \\
\midrule
\multirow{3}{*}{CACAPO (nl)} & No\_Ext       & 14.65 & \textbf{7.70} & 2748  & 0.58 & 0.85 & 98.86  & 0.56 & 0.20 & 0.52 \\
 & Dat\_Aug    & 14.31 & 6.09 & 2828  & 0.63 & \textbf{0.91} & \textbf{98.93}  & 0.56 & 0.22 & 0.57 \\
 & Pseu\_Lab    & \textbf{14.99} & 6.30 & \textbf{3176}  & \textbf{0.65} & \textbf{0.91} & 93.54  & \textbf{0.64} & \textbf{0.24} & \textbf{0.66} \\
\midrule
\multirow{3}{*}{E2E} & No\_Ext       & 28.58 & 7.66 & 120   & 0.34 & 0.50 & \textbf{100} & 0.11 & 0.00 & \textbf{0.11} \\
 & Dat\_Aug    & \textbf{34.42} & \textbf{7.73} & \textbf{223}   & \textbf{0.38} & \textbf{0.55} & \textbf{100} & \textbf{0.16} & \textbf{0.03} & 0.10 \\
 & Pseu\_Lab    & 23.22 & 5.26 & 115   & 0.26 & 0.38 & \textbf{100} & 0.07 & \textbf{0.03} & 0.08 \\
\midrule
\multirow{3}{*}{WebNLG} & No\_Ext       & 16.01 & 6.71 & 2136  & 0.43 & 0.68 & 79.78  & 0.69 & 0.02 & \textbf{0.69} \\
 & Dat\_Aug    & 15.87 & \textbf{6.91} & 2311  & 0.43 & 0.71 & \textbf{97.71}  & 0.62 & \textbf{0.15} & 0.49 \\
 & Pseu\_Lab    & \textbf{16.42} & 6.70 & \textbf{2404}  & \textbf{0.45} & \textbf{0.72} & 81.13  & \textbf{0.73} & 0.08 & 0.66 \\
\bottomrule
\end{tabular*}
\caption{Average sentence length, standard deviation of sentence length, mean-segmented type-token ratio (TTR), bigram TTR, percentage novel descriptions, coverage, novelty and local recall with importance class 1 (bold = highest) per dataset and (XL-format) semi-supervised learning approach.}
\label{tab:divmetsperdataset}
\end{table}

\begin{figure}
\begin{minipage}{.45\textwidth}
  \includegraphics[width=1\linewidth]{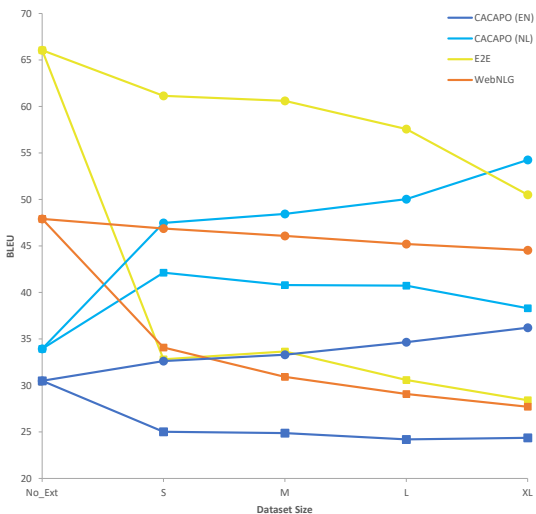}
  \caption{\label{BLEU_Fig} BLEU scores of the datasets per dataset extension. Round markers = pseudo-labeling; square markers = data augmentation.}
\end{minipage}%
\begin{minipage}{.1\textwidth}
\end{minipage}
\begin{minipage}{.45\textwidth}
  \includegraphics[width=1\linewidth]{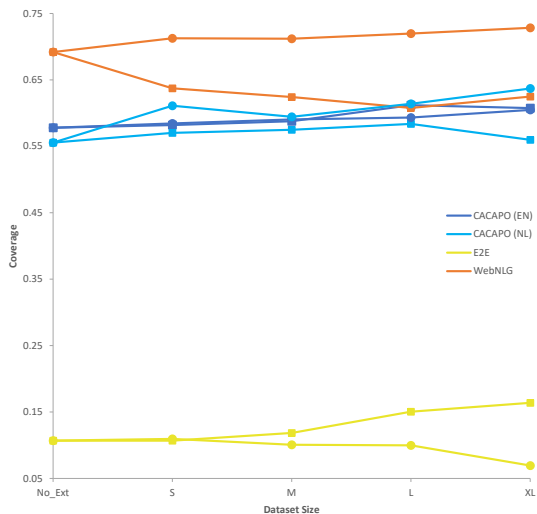}
  \caption{\label{Cov_Fig} Coverage scores of the datasets per dataset extension. Round markers = pseudo-labeling; square markers = data augmentation.}
\end{minipage}
\end{figure}

A summary of the diversity metrics can be found in \hyperref[tab:divmetsperdataset]{Table~\ref*{tab:divmetsperdataset}} on a dataset level, and on domain level in \hyperref[tab:divmetsslm]{Table~\ref*{tab:divmetsslm}}. Furthermore, \hyperref[Cov_Fig]{Figure~\ref*{Cov_Fig}} shows the impact of an increase in the synthetic data from the semi-supervised learning methods on coverage scores. Overall, data extensions generally lead to higher diversity scores. The semi-supervised learning approaches seem to generate more diverse output compared to no training set extension, but the semi-supervised learning approach that gives the most diverse output seems to differ per dataset (domain) and per diversity metric. Therefore, we will discuss the outcomes grouped by different diversity metrics.

Average sentence length and standard deviation of the sentence can be an indicator of perceived diversity in a text: longer sentences tend to contain more variation, and bigger differences between sentence length makes the output more heterogeneous. It could be expected that especially the pseudo-labeling approach affects sentence length standard deviation if the sentences that this approach introduces are also of a varied sentence length, while the average sentence length should not change too much for CACAPO and E2E, as the newly introduced sentences come from similar sources as the sentences in the original training set. For WebNLG, changes in average sentence length can be expected for the pseudo-labeling approach as the sentences are from a different source (Wikipedia vs. Crowdsourced). The data augmentation approach is expected to keep the standard deviation of sentence length, and average sentence length similar, as this approach perturbs words but generally keeps sentence structure and length the same. The effect of pseudo-labeling, however is only partially according to expectations; pseudo-labeling obtained the highest standard deviation score for 9 of the 19 domains, but only for 1 dataset overall (see \hyperref[tab:divmetsperdataset]{Table~\ref*{tab:divmetsperdataset}}; \hyperref[tab:divmetsslm]{Table~\ref*{tab:divmetsslm}}). In terms of sentence length, it only shows a clear difference (decrease) compared to the no extension approach for E2E, and only marginal differences for the other datasets and dataset domains. Data augmentation indeed shows similar scores for average sentence length and standard deviation of sentence length consistently amongst datasets and dataset domains.

The number of types, type-token ratios, percentage of novel texts, and the novelty score are all direct indicators of lexical diversity. For these metrics, the improvements of the semi-supervised learning approaches are also the most pronounced. The data augmentation and pseudo-labeling approach each seemed to perform best in terms of increasing lexical diversity on roughly half of the datasets and dataset domains. 
For the English CACAPO, and E2E, data augmentation seemed to result in the highest lexical diversity scores, while these scores were highest for the pseudo-labeling approach in the case of Dutch CACAPO and WebNLG. For the Dutch CACAPO dataset, this may have to do with the nature of the language model used for data augmentation: BERT-large for English \citep{devlin_etal_2019} is likely better able to inject diverse perturbations compared to BERTje \citep{devries_etal_2019} which is a Dutch translation of BERT-base. For WebNLG it might have to do with the nature of the texts that were used for pseudo-labeling: the Wikipedia texts are probably more dissimilar (thus injecting more diversity) to the texts in the training set, compared to the pseudo-labeled texts used for the other datasets. It is worth noting that the novelty scores of the no extension approach for the WebNLG and E2E datasets are consistently close to 0, meaning that (almost) no new words have been introduced for these datasets that were not found in the training or development data. This absence of new words possibly showcases catastrophic forgetting \citep{greco_etal_2019}, which may occur when a training set is decently sized for finetuning, which leads to a neural NLG system forgetting about the language model's capabilities. Also worth noting is that the percentage of novel texts shows that (almost) all generated texts in CACAPO and E2E are different from the texts in the training data, thus creating a ceiling effect for these datasets. This has to do with the setup of the test set, where input data rarely, if ever, overlaps with the input data in the training and development sets. This metric is more interesting for WebNLG, where we see that non-novel sentences are only rarely generated with the data augmentation approach, while this does occur more frequently with the pseudo-labeling approach and no extension approach.

Coverage and local recall measure text retention in the output compared to the training set. Coverage is a global recall metric, measuring how many of the word types from the training and development sets were retained in the output. A higher coverage score suggests that more of the naturally occurring variation in the training and development set is retained in the test set. Local recall compares the content words in the generated output to those in the reference in the test set. A higher score on lexical variation metrics, paired with a high score on local recall indicates that higher diversity was not obtained at the cost of meaningful content words. Regarding coverage, we again see that the semi-supervised learning approaches consistently outperform the system that only used the standard training set. The highest coverage scores for WebNLG and Dutch CACAPO were achieved with pseudo-labeling, while data augmentation achieved the highest coverage score for English CACAPO and for E2E. Furthermore, \hyperref[Cov_Fig]{Figure~\ref*{Cov_Fig}} shows that coverage scores are generally increased when the training set size is increased with synthetic data. This effect is most pronounced for pseudo-labeling, whereas an increase in training set size does not necessarily lead to higher coverage scores with data augmentation; while training size increases with data augmentation do seem to have a positive effect on coverage for CACAPO, they seem to decrease coverage scores for WebNLG and E2E. For local recall, we see that the pseudo-labeling approach achieved the highest scores for Dutch and English CACAPO. The small-scale nature of these domains might stand to benefit more from the extra training data that the pseudo-labeling approach offers to retain commonly occurring content words. For WebNLG and E2E we see that the system trained on the original dataset retains the most content words, while the pseudo-labeling approach generally stays relatively close in terms of local recall scores. The data augmentation approach shows a more sizable drop-off, which makes sense as the content words were augmented for this semi-supervised learning approach.

In sum, in support of H2, we find that semi-supervised learning increases the output diversity. Data augmentation seems to be the best method to achieve this for English CACAPO and E2E, whereas the pseudo-labeling approach achieves better diversity scores for Dutch CACAPO and WebNLG. The results also indicate that the neural NLG system without an semi-supervised learning approach suffers from catastrophic forgetting: overfitting on the training set too tightly which means a relative lack of output diversity as a result. We do not find a marked difference between datasets in terms of the effects of semi-supervised learning in general on output diversity. All datasets seem to benefit relatively equally from semi-supervised learning. Therefore, no clear effect of dataset type (crowdsourcing based vs. naturally occurring) and language (Dutch versus English) was detected, which is contrary to H4 and H5 respectively.

\begin{figure}
\begin{minipage}{.5\textwidth}
  \includegraphics[width=1\linewidth]{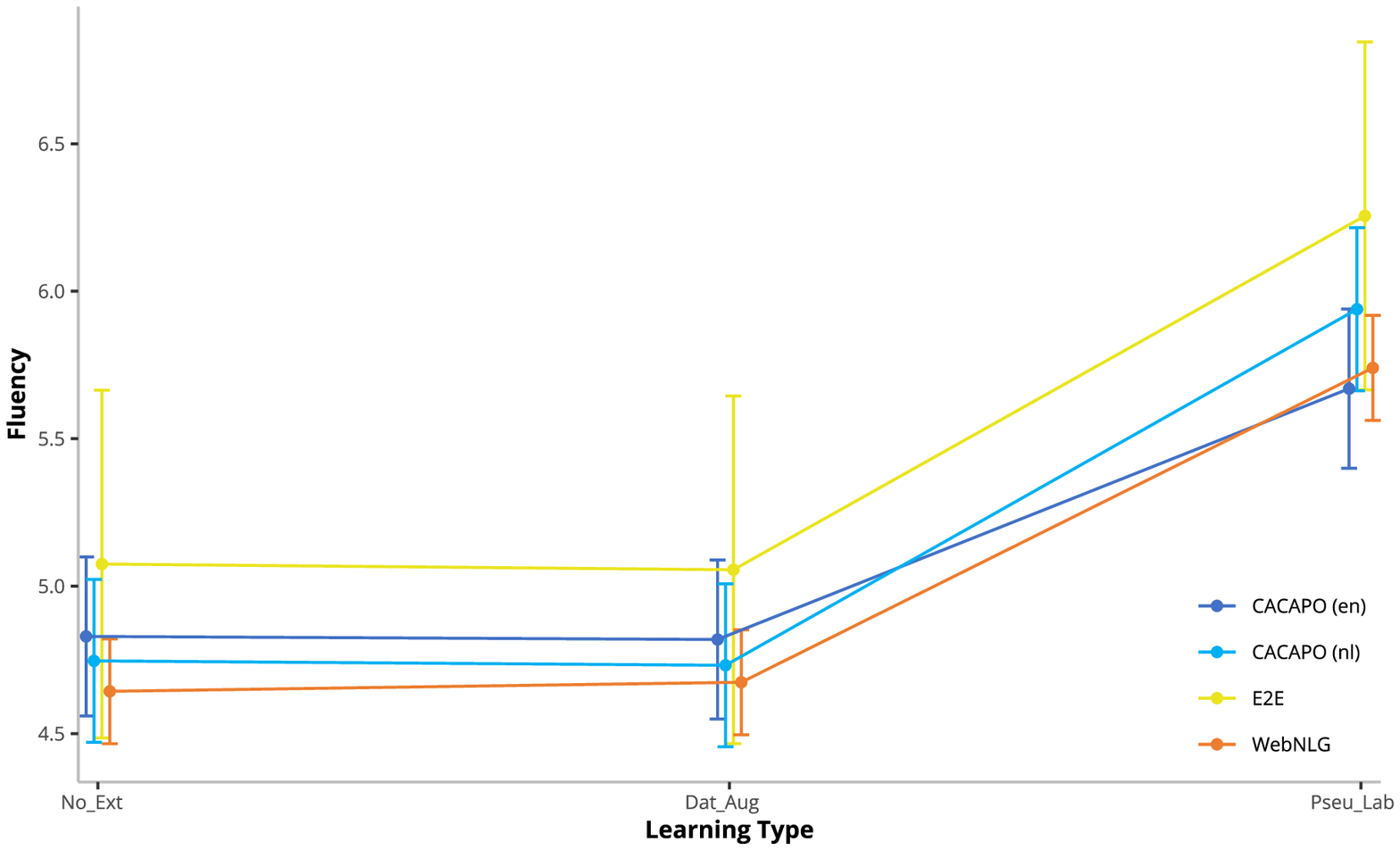}
  \caption{\label{FluenFig} Mean fluency per dataset and learning type.}
\end{minipage}%
\begin{minipage}{.5\textwidth}
  \includegraphics[width=1\linewidth]{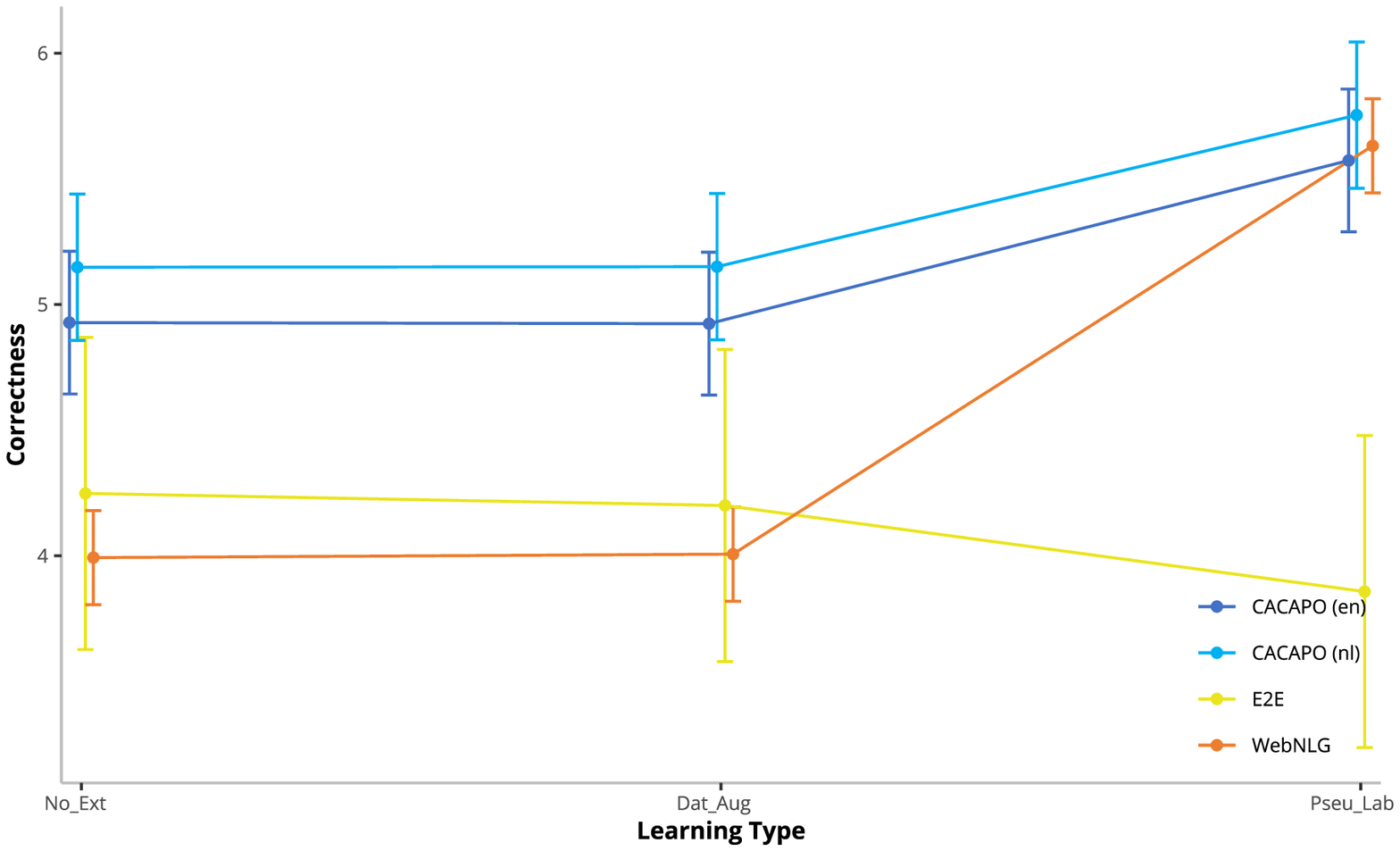}
  \caption{\label{CorrFig} Mean correctness per dataset and learning type.}
\end{minipage}
\centering
\begin{minipage}{.5\textwidth}
\includegraphics[width=1\textwidth]{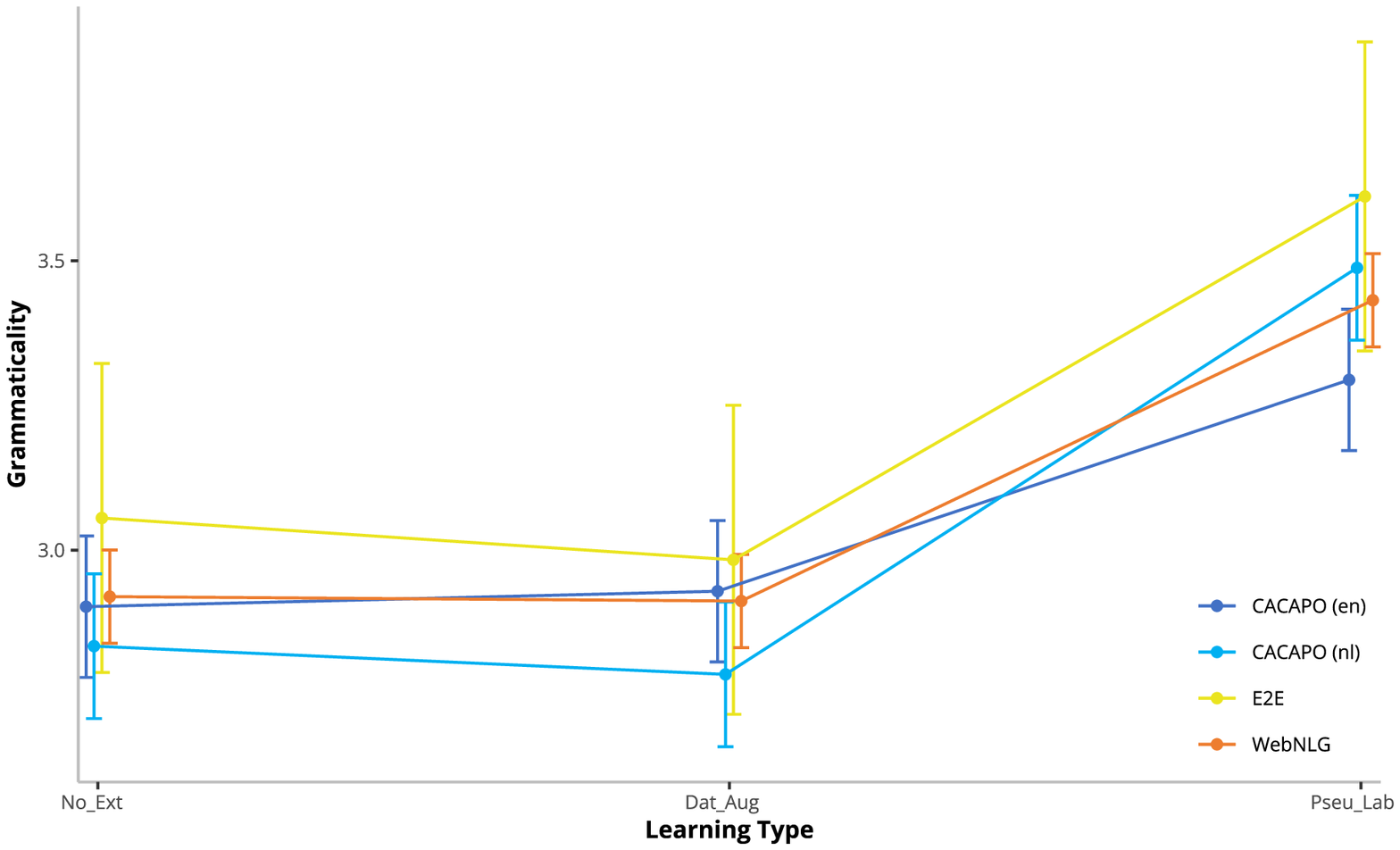} \\
\caption{\label{GramFig} Mean grammaticality per dataset and learning type.}
\end{minipage}
\end{figure}


\subsection{Quantitative Human Evaluation}

In addition to the automatic evaluations, we also conducted a human evaluation to investigate perceived quality, by measuring fluency, correctness and grammaticality. The overall scores are summarized in \hyperref[tab:meanhumevalperdataset]{Table~\ref*{tab:meanhumevalperdataset}}, and reported per sub-corpus in \hyperref[tab:meanhumevalperdomain]{Table~\ref*{tab:meanhumevalperdomain}}. Inspection of \hyperref[tab:meanhumevalperdataset]{Table~\ref*{tab:meanhumevalperdataset}} reveals a clear pattern: in almost all cases the highest scores are obtained with the pseudo-labeling approach. This was further examined by conducting a series of Linear Mixed Models, with semi-supervised methods and the datasets/dataset domains as independent variables and fluency, correctness, and grammaticality as the dependent variables. Linear Mixed Models enable us to control for the systematic variation due to the nested/clustered structure of the data caused by participants having rated multiple texts. Furthermore, linear mixed models are able to deal appropriately with unequal sample sizes (i.e., language and dataset in this experiment). To do so, we added dataset, domain and participant as a nested random factor in the models. Following \cite{vanmiltenburg_etal_2019}, we used the \texttt{lme4} \citep{bates_etal_2015} package in \texttt{R}, to build our Linear Mixed Models using the \texttt{lmer} function and estimate p-values for the models, respectively. Datasets and dataset domains were investigated using seperate models, and a separate model for each dependent variable was also necessary, meaning that a total of 6 models were created. Furthermore, ``no extension'' served as the reference level intercept for semi-supervised learning approach, and ``E2E'' as the reference level intercept for dataset and dataset domain. The \texttt{emmeans} package was used for pairwise comparisons when analyzing the differences between semi-supervised learning approaches, datasets, dataset domains, and interaction effects if the Linear Mixed Model showed significant main effects or interaction effects. All models converged and the visual check of model assumptions using the residual plots indicated no signs of violations.

\subsubsection*{Fluency}

The model for fluency was significant (conditional R$^2$ = 0.23, marginal R$^2$ = 0.07, \textit{p} < .001). The model showed a main effect of training type, but not of dataset, and also did not suggest any interaction effects. \hyperref[tab:meanhumevalperdataset]{Table~\ref*{tab:meanhumevalperdataset}} represents the mean scores and training type differences, while the mixed model results are summarized in \hyperref[mmstats]{Table~\ref*{mmstats}}. Significant effects were further investigated using estimated marginal means with Bonferroni correction, which corroborated these findings. Only the pairwise comparisons for semi-supervised learning approaches showed significance: there was no significant difference in fluency between no extension (\textit{M} = 4.73, \textit{SD} = 2.05) and data augmentation (\textit{M} = 4.74 \textit{SD} = 2.06), but the fluency scores for the pseudo-labeling approach (\textit{M} = 5.79 \textit{SD} = 1.62) were significantly higher than both the data augmentation approach and no extension. This effect was robust across all dataset domains as well, with all but 4 domains showing the same pattern between semi-supervised learning approaches (see \hyperref[tab:meanhumevalperdomain]{Table~\ref*{tab:meanhumevalperdomain}}).

\hyperref[FluenFig]{Figure~\ref*{FluenFig}} 
illustrate these results: all datasets show a similar pattern, where the fluency scores between no extension and data augmentation are virtually similar, and the perceived fluency increases for pseudo-labeling. Sentences were thus perceived as more fluent when a training set was enriched with data created via the pseudo-labeling approach. Furthermore, perceived fluency scores were relatively similar for all the investigated datasets, and the fluency differences between the semi-supervised learning approaches were similar across all datasets. 

Partly in support of H1, semi-supervised learning seems to increase the output quality compared to a language model only trained on the base training set. However, this is only the case for the pseudo-labeling approach and not for the data augmentation approach (RQ1). Furthermore, these fluency results do not support H3, nor H5: the beneficial effect of semi-supervised learning is consistent regardless of whether the original dataset is small-scale or large-scale (H3), Dutch or English (H5).

\begin{table}[t]
\centering
\footnotesize
\begin{tabular*}{\textwidth}{l @{\extracolsep{\fill}} lrrrr}
\toprule
\bf Dataset     & \bf Train Type & \bf N  & \bf Fluency     & \bf Correctness & \bf Grammaticality \\
\midrule
\multirow{3}{*}{CACAPO (en)} & No\_Ext    & \multirow{3}{*}{43} & 4.83 (2.05)$^a$ & 4.93 (1.86)$^a$ & 2.90 (0.98)$^a$    \\
 & Dat\_Aug   & & 4.82 (2.05)$^a$ & 4.92 (1.89)$^a$ & 2.93 (0.98)$^a$    \\
 & Pseu\_Lab  & & \textbf{5.67 (1.70)$^b$} & \textbf{5.57 (1.70)$^b$} & \textbf{3.29 (0.86)$^b$}    \\
\midrule
\multirow{3}{*}{CACAPO (nl)} & No\_Ext    & \multirow{3}{*}{41} & 4.75 (2.03)$^a$ & 5.15 (1.90)$^a$ & 2.83 (1.09)$^a$    \\
 & Dat\_Aug   & & 4.73 (2.09)$^a$ & 5.15 (1.88)$^a$ & 2.79 (1.12)$^a$    \\
 & Pseu\_Lab  & & \textbf{5.94 (1.52)$^b$} & \textbf{5.75 (1.58)$^b$} & \textbf{3.49 (0.78)$^b$}    \\
\midrule
\multirow{3}{*}{WebNLG} & No\_Ext    & \multirow{3}{*}{99} & 4.64 (2.08)$^a$ & 3.99 (2.21)$^a$ & 2.92 (1.02)$^a$    \\
 & Dat\_Aug   & & 4.67 (2.08)$^a$ & 4.01 (2.22)$^a$ & 2.91 (1.02)$^a$    \\
 & Pseu\_Lab  & & \textbf{5.74 (1.66)$^b$} & \textbf{5.63 (1.82)$^b$} & \textbf{3.43 (0.79)$^b$}    \\
\midrule
\multirow{3}{*}{E2E} & No\_Ext    & \multirow{3}{*}{9} & 5.08 (1.73)$^a$ & \textbf{4.25 (1.86)$^a$} & 3.06 (0.88)$^a$    \\
 & Dat\_Aug   & & 5.06 (1.70)$^a$ & 4.20 (1.83)$^a$ & 2.98 (0.91)$^a$    \\
 & Pseu\_Lab  & & \textbf{6.26 (1.02)$^b$} & 3.86 (1.71)$^a$ & \textbf{3.61 (0.65)$^b$}   \\
\bottomrule
\end{tabular*}
\caption{Mean fluency, correctness, and grammaticality per semi-supervised learning type for each dataset (SDs between brackets). Different superscripts indicate significant differences between semi-supervised learning approaches. Higher scores mean more positively perceived output.}
\label{tab:meanhumevalperdataset}
\end{table}

\subsubsection*{Correctness}
For correctness, the model was significant as well (conditional R$^2$ = 0.29, marginal R$^2$ = 0.11, \textit{p} < .001). The model showed a main effect of semi-supervised learning approach, dataset, and an interaction between the two. \hyperref[tab:meanhumevalperdataset]{Table~\ref*{tab:meanhumevalperdataset}} shows the mean scores and semi-supervised learning approach differences, and \hyperref[mmstats]{Table~\ref*{mmstats}} a summary of the mixed models for correctness. The significant effects were further explored using estimated marginal means with Bonferroni correction. For semi-supervised learning type, there was no significant difference between no extension (\textit{M} = 4.46 \textit{SD} = 2.12) and the data augmentation approach (\textit{M} = 4.47 \textit{SD} = 2.12), but there were significantly higher correctness scores for the pseudo-labeling approach (\textit{M} = 5.56 \textit{SD} = 1.78) compared to both other approaches. For dataset, there was no significant difference between E2E (\textit{M} = 4.10 \textit{SD} = 1.80) and WebNLG (\textit{M} = 4.54 \textit{SD} = 2.23), and no difference between English CACAPO (\textit{M} = 5.14 \textit{SD} = 1.84) and Dutch CACAPO (\textit{M} = 5.35 \textit{SD} = 1.81). However, both CACAPO datasets scored significantly higher on perceived correctness than both WebNLG and E2E. On a dataset domain level, especially the \texttt{Astronaut}, \texttt{Food}, and \texttt{Monument} domains of WebNLG performed worse for WebNLG. 

The interaction effect can be explained in more detail using \hyperref[CorrFig]{Figure~\ref*{CorrFig}} 
as well as the simple effects analysis. Regarding differences between datasets, differentiated per semi-supervised learning approach, we see that for no extension, WebNLG (\textit{M} = 3.99, \textit{SD} = 2.21) obtained significantly worse correctness scores compared to both Dutch CACAPO (\textit{M} = 5.15, \textit{SD} = 1.90) and English CACAPO (\textit{M} = 4.93, \textit{SD} = 1.86). This pattern is similar for data augmentation: WebNLG (\textit{M} = 4.01, \textit{SD} = 2.22) performed significantly worse in terms of correctness compared to both Dutch CACAPO (\textit{M} = 5.15, \textit{SD} = 1.88) and English CACAPO (\textit{M} = 4.92, \textit{SD} = 1.89). This consistency also makes sense when comparing the datasets' scores on no extension to data augmentation: none of the datasets' scores on data augmentation differ significantly from the scores they obtained on no extension. However, the pattern is different for pseudo-labeling: the correctness scores for English CACAPO (\textit{M} = 5.57, \textit{SD} = 1.70), Dutch CACAPO (\textit{M} = 5.75, \textit{SD} = 1.58), and WebNLG (\textit{M} = 5.63, \textit{SD} = 1.82) are significantly higher for pseudo-labeling than they are for both no extension and data augmentation, resulting in no significant differences in correctness scores between the three datasets for pseudo-labeling. However, the correctness scores for E2E on pseudo-labeling (\textit{M} = 3.86, \textit{SD} = 1.71) is not significantly different compared to the correctness scores E2E obtained on no extension and data augmentation. This results in E2E performing significantly worse on pseudo-labeling compared to English CACAPO, Dutch CACAPO, and WebNLG. On a domain level, we see that the differences between semi-supervised learning approaches are most pronounced for the WebNLG domains, while the differences are not as clear for the English CACAPO and Dutch CACAPO domains (see \hyperref[tab:meanhumevalperdomain]{Table~\ref*{tab:meanhumevalperdomain}}).

In sum, the data shows that overall, no extension and data augmentation performed similarly on perceived correctness across all datasets, and that pseudo-labeling also leads to the highest correctness scores as it does for fluency scores. However, this effect is moderated by the dataset. E2E did not achieve higher correctness scores with the pseudo-labeling approach compared to the other approaches, while the other datasets did. Similar to the fluency results, the correctness scores are partially in support of H1, showing that the semi-supervised learning shows higher correctness scores compared to a language model only trained on the base training set, but only in the case of the pseudo-labeling approach and not for the data augmentation approach (RQ1). The correctness scores are partially in support of H3, as the beneficial effect of semi-supervised learning could not be found for the largest-scale dataset (E2E), but it did for WebNLG. Finally, the correctness scores do not seem to differ per language, which does not support H5.

\subsubsection*{Grammaticality}
The model for grammaticality was significant (conditional R$^2$ = 0.20, marginal R$^2$ = 0.07, \textit{p} < .001). Similar to the fluency model, the grammaticality model showed a main effect of training type, but not of dataset, and also did not suggest any interaction effects. Mean scores for grammaticality and differences for training type are shown in \hyperref[tab:meanhumevalperdataset]{Table~\ref*{tab:meanhumevalperdataset}} and the mixed model results are summarized in \hyperref[mmstats]{Table~\ref*{mmstats}}. The estimated marginal means with Bonferroni correction shows that no extension (\textit{M} = 2.90, \textit{SD} = 1.02) was not significantly different compared to data augmentation (\textit{M} = 2.89, \textit{SD} = 1.03) in terms of grammaticality, and that both no extension and data augmentation obtained significantly worse grammaticality scores compared to pseudo-labeling (\textit{M} = 3.42, \textit{SD} = 0.80). The differences between semi-supervised learning approaches are also fairly stable on a category level, with 14 out of 19 domains having no significant difference between no extension and data augmentation, and pseudo-labeling performing significantly better than the other two approaches.

\hyperref[GramFig]{Figure~\ref*{GramFig}} 
shows a similar pattern as was found for fluency. The pseudo-labeling approach led to higher grammaticality scores than no extension and data augmentation did. Furthermore, the effects of grammaticality seem to be consistent amongst datasets: the datasets did not differ significantly from each other in terms of the obtained grammaticality scores, and performed similarly with the semi-supervised learning approaches.

\subsection{Error Analysis}

\begin{table*}
\centering
\setlength{\tabcolsep}{1.8pt}
\renewcommand{\arraystretch}{0.95}
\scriptsize
\begin{minipage}{.49\linewidth}
\centering
\begin{tabular}{lrrrr}
\toprule 
\bf CACAPO (NL) & \bf NoExt & \bf DatAug & \bf PseuLab & \bf Total \\
\midrule
Hallucinations & 6$^a$ & 7$^a$ & 5$^a$ & 18  \\
Missing info & 4$^a$ & 3$^a$ & 3$^a$ & 10 \\
References & 2$^a$ & 3$^a$ & 1$^a$ & 6 \\
Verb form & 4$^a$ & 2$^a$ & 1$^a$ & 7 \\
Determiners & 2$^a$ & 1$^a$ & 0$^a$ & 3 \\
Punct./capital. & 0$^a$ & 0$^a$ & 0$^a$ & 0 \\
Lexical choices & 4$^a$ & 4$^a$ & 2$^a$ & 10 \\
Repetition & 3$^a$ & 3$^a$ & 1$^a$ & 7 \\
Connections & 7$^a$ & 4$^a$ & 1$^a$ & 12 \\
Missing parts & 4$^a$ & 2$^a$ & 2$^a$ & 8 \\
Other & 2$^a$ & 3$^a$ & 2$^a$ & 7 \\
\midrule
Total & 38 & 32 & 18 & 88 \\
\bottomrule
\end{tabular}
\end{minipage}
\begin{minipage}{.49\linewidth}
\centering
\begin{tabular}{lrrrr}
\toprule 
\bf CACAPO (EN) & \bf NoExt & \bf DatAug & \bf PseuLab & \bf Total \\
\midrule
Hallucinations & 9$^a$ & 10$^a$ & 5$^a$ & 24 \\
Missing info & 2$^a$ & 0$^a$ & 0$^a$ & 2 \\
References & 0$^a$ & 0$^a$ & 1$^a$ & 1 \\
Verb form & 2$^a$ & 2$^a$ & 1$^a$ & 5 \\
Determiners & 0$^a$ & 1$^a$ & 0$^a$ & 1 \\
Punct./capital. & 2$^a$ & 1$^a$ & 1$^a$ & 4 \\
Lexical choices & 3$^a$ & 4$^a$ & 0$^a$ & 7 \\
Repetition & 2$^a$ & 5$^a$ & 0$^a$ & 7 \\
Connections & 2$^a$ & 4$^a$ & 1$^a$ & 7 \\
Missing parts & 5$^a$ & 5$^a$ & 2$^a$ & 12 \\
Other & 3$^a$ & 5$^a$ & 5$^a$ & 13 \\
\midrule
Total & 30 & 37 & 16 & 83 \\
\bottomrule
\end{tabular}
\end{minipage}
\vspace{5pt}
\begin{minipage}{.49\linewidth}
\centering
\begin{tabular}{lrrrr}
\toprule 
\bf WebNLG \hspace{23.5pt} & \bf NoExt & \bf DatAug & \bf PseuLab & \bf Total \\
\midrule
Hallucinations & 5$^a$ & 5$^a$ & 5$^a$ & 15 \\
Missing info & 11$^a$ & 12$^a$ & 6$^a$ & 29 \\
References & 3$^a$ & 7$^a$ & 4$^a$ & 14 \\
Verb form & 1$^a$ & 2$^a$ & 1$^a$ & 4 \\
Determiners & 1$^a$ & 2$^a$ & 1$^a$ & 4 \\
Punct./capital. & 1$^a$ & 0$^a$ & 0$^a$ & 1 \\
Lexical choices & 2$^a$ & 1$^a$ & 0$^a$ & 3 \\
Repetition & 8$^a$ & 7$^a$ & 0$^a$ & 15 \\
Connections & 2$^a$ & 3$^a$ & 3$^a$ & 8 \\
Missing parts & 6$^a$ & 9$^a$ & 2$^a$ & 17 \\
Other & 2$^{a,b}$ & 0$^b$ & 4$^a$ & 6 \\
\midrule
Total & 42 & 48 & 26 & 116 \\
\bottomrule
\end{tabular}
\end{minipage}
\begin{minipage}{.49\linewidth}
\centering
\begin{tabular}{lrrrr}
\toprule 
\bf E2E \hspace{46.5pt} & \bf NoExt & \bf DatAug & \bf PseuLab & \bf Total \\
\midrule
Hallucinations & 9$^a$ & 10$^a$ & 13$^a$ & 32 \\
Missing info & 5$^{a,b}$ & 3$^b$ & 10$^a$ & 18 \\
References & 8$^a$ & 7$^a$ & 4$^a$ & 19 \\
Verb form & 1$^a$ & 0$^a$ & 0$^a$ & 1 \\
Determiners & 2$^a$ & 1$^a$ & 0$^a$ & 3 \\
Punct./capital. & 2$^a$ & 4$^a$ & 0$^a$ & 6 \\
Lexical choices & 1$^a$ & 0$^a$ & 0$^a$ & 1 \\
Repetition & 4$^a$ & 5$^a$ & 0$^a$ & 9 \\
Connections & 0$^a$ & 0$^a$ & 0$^a$ & 0 \\
Missing parts & 1$^a$ & 2$^a$ & 0$^a$ & 3 \\
Other & 2$^a$ & 5$^a$ & 0$^a$ & 7 \\
\midrule
Total & 35 & 37 & 27 \\
\bottomrule
\end{tabular}
\end{minipage}
\begin{tabular}{lrrrr}
\toprule 
\bf All Datasets & \bf NoExt & \bf DatAug & \bf PseuLab & \bf Total \\
\midrule
Hallucinations & 29$^a$ & 32$^a$ & 28$^a$ & 89 \\
Missing info & 22$^a$ & 18$^a$ & 19$^a$ & 59 \\
References & 13$^a$ & 17$^a$ & 10$^a$ & 40 \\
Verb form & 8$^a$ & 6$^a$ & 3$^a$ & 17 \\
Determiners & 5$^a$ & 5$^a$ & 1$^a$ & 11 \\
Punct./capital. & 5$^a$ & 5$^a$ & 1$^a$ & 11 \\
Lexical choices & 10$^a$ & 9$^a$ & 2$^a$ & 21 \\
Repetition & 17$^a$ & 20$^a$ & 1$^b$ & 38 \\
Connections & 11$^a$ & 11$^a$ & 5$^a$ & 27 \\
Missing parts & 16$^a$ & 18$^a$ & 6$^a$ & 40 \\
Other & 9$^a$ & 13$^a$ & 11$^a$ & 33 \\
\midrule
Total & 145 & 154 & 87 & 386 \\
\bottomrule
\end{tabular}
\caption{Number of errors for each dataset per error category and semi-supervised learning approach. Different superscripts indicate significant differences between semi-supervised learning approaches.}
\label{tab:errornumber}
\end{table*}

The goal of the error analysis was to investigate whether the different systems were prone to other kinds of errors, which could indicate points of attention for future work. Hence, a chi-square test was conducted with error type (see \hyperref[erroranalysismethod]{Section~\ref*{erroranalysismethod}}), semi-supervised learning approach (no extension, data augmentation, and pseudo-labeling), and dataset (Dutch CACAPO, English CACAPO, WebNLG, and E2E) as categorical variables. Pairwise comparisons between systems were made using Bonferroni adjusted z-tests for column proportions. Results of the comparisons are shown in \hyperref[tab:errornumber]{Table~\ref*{tab:errornumber}}.

The chi-square test did not show a significant difference in error proportions between the semi-supervised learning approaches for Dutch CACAPO ($\chi^2$(18) = 6.45, \textit{p} = .994), English CACAPO ($\chi^2$(20) = 17.82, \textit{p} = .599), WebNLG ($\chi^2$(20) = 21.11, \textit{p} = .391), E2E ($\chi^2$(18) = 28.52, \textit{p} = .055), as well as all the datasets combined ($\chi^2$(20) = 26.43, \textit{p} = .152). This could indicate that differences found in the quantitative human analysis are generally not the result of distinct categorical error differences, but rather an overall difference in all errors combined. This is corroborated by the total number of errors for the three systems: the pseudo-labeling approach yielded only around half of the errors of the other two approaches in total, and this effect was consistent for every dataset (see \hyperref[tab:errornumber]{Table~\ref*{tab:errornumber}}). 

Despite no overall effect being found, the pairwise comparisons do show a few significant differences. For WebNLG, a larger proportion of \textit{Other} errors were found for the pseudo-labeling approach (15.4\%) compared to the data augmentation approach (0.0\%), but this difference is likely explained by the small number of error observations for this particular error category and dataset. More interesting is the difference found for the E2E dataset regarding the number of missing information errors obtained by the data augmentation approach (8.1\%), compared to the pseudo-labeling approach (37\%), which could be the reason why the correctness score of the pseudo-labeling approach was lower for the E2E dataset compared to the other datasets in the quantitative human evaluation study. Furthermore, the errors of all datasets combined showed that the pseudo-labeling approach had much less issues with the repetition of words/phrases (1.1\% of errors for pseudo-labeling, compared to 11.7\% and 13.0\% for no extension and data augmentation, respectively). These issues are seen in sentences such as ``\textit{Batchoy is eaten in the their, a the of the their spoken is is.}'' and may indicate an issue of undertraining, which is solvable by extending the training set via the pseudo-labeling approach. See \hyperref[tab:errornumber]{Table~\ref*{tab:errornumber}}.

\section{Discussion and Conclusion}
This study investigated the potentially beneficial effect of semi-supervised learning in combination with a language model. More specifically, it investigated whether enriching a training set via the data augmentation approach (i.e., generating several variants of a training text by replacing certain words with synonyms and semantically similar words) and a pseudo-labeling approach (i.e., labeling unlabeled texts using an information extraction model trained on the existing labeled training data) could increase the performance of data-to-text NLG that already utilizes a large-scale language model (T5-large/mT5-large). Previous work has found that semi-supervised learning could increase the output quality of an NLG system \citep[e.g.,][]{qader_etal_2019,schmitt_etal_2020,su_etal_2020,tseng_etal_2020,chang_etal_2021} and that utilizing language models for data-to-text NLG could help improve output quality as well \citep[e.g.,][]{kale_rastogi_2020}. However, not much is known about the combination of language models and semi-supervised learning in a data-to-text generation setting. Therefore, it is not known whether semi-supervised approaches are still effective when language models are also used, and if they are: under what conditions they are effective. Besides the type of semi-supervised learning as a condition, this study also investigated multiple datasets with different characteristics to see whether they affected the effectiveness of semi-supervised learning in combination with language models. 

\subsubsection*{Output Quality}
Partial support was found for the hypothesis that semi-supervised learning positively affects output quality (H1). When observing the results of the quantitative human evaluation, the pseudo-labeling approach consistently obtained higher scores on correctness, grammaticality, and fluency compared to the data-to-text system with a language model that was only trained on the base training set (hereafter: no extension approach). This is in line with \cite{sun_etal_2020} who found for a text classification task that a combination of a language model and pseudo-labeling led to the highest scores. It is also corresponding with previous research that used joint learning systems for data-to-text generation \citep[e.g.,][]{qader_etal_2019,schmitt_etal_2020,su_etal_2020,tseng_etal_2020,chang_etal_2021}, with the difference that this research compared it to a system where a language model was finetuned on the training set, rather than a system that was trained on merely the training set. 

However, the data augmentation approach performed equal to the no extension approach in the quantitative human evaluation study and error analysis, and the output quality scores were generally worse than the quality scores yielded by the pseudo-labeling approach. These results suggest that the pseudo-labeling approach is a better semi-supervised learning approach than the data augmentation approach if the goal is to increase the output quality over a system that is only trained on the base training set (RQ1). These results make intuitive sense: the data augmentation approach only makes subtle changes to the data that is represented in the text, and does not make any fundamental adjustments to the sentence structure compared to the original text, thus keeping the output relatively similar to that of the no extension approach. The pseudo-labeling approach introduces not only new sentence structures to the training set, but also a large amount of new data types and combinations of data. This increased variety in training data might lead to better generalizations and better handling of the diverse situations in the test set, subsequently leading to higher quality output and less undertraining issues (as was corroborated by the error analysis). Thus, while previous research with different data augmentation approaches found data augmentation to increase performance compared to a system that was only trained on the training set \citep[e.g.,][]{kulhanek_etal_2021,riabi_etal_2021,tandon_etal_2018,alberti_etal_2019, chang_etal_2021b,kedzie_mckeown_2019}, it may not result in text quality improvements if a language model is also implemented. It is possible that when the original training set is too small (which is common for NLG datasets), it may lead to an undertrained system. Undertraining can be reduced by implementing a language model that is finetuned on the training data, and by extending the training data with data augmentation. However, this combination of the two might be redundant for handling the undertraining issue.

The automatic metrics for output quality estimation suggest that the pseudo-labeling approach is the most effective approach for the CACAPO dataset---especially the Dutch CACAPO dataset---and that the no extension approach leads to the best output for E2E, and WebNLG (in concordance with H3 which states that smaller-scale datasets benefit more from semi-supervised learning, and H5 that states that Dutch datasets benefit more than English datasets). However, the results of the quantitative human evaluation shows that the fluency and grammaticality outcomes are fairly consistent amongst datasets. The exception being correctness, where semi-supervised learning did not return higher ratings than the no extension approach for E2E. However, it is possible that this has more to do with the nature of the extensions used with E2E for the pseudo-labeling approach, as there was a clear correctness increase for the pseudo-labeling approach over the no extension approach for WebNLG. The semantic parser that was being applied on the extra texts for the pseudo-labeling approach seemed to be struggling (with F1 scores of only around 57\% on the additional data), possibly due to overfitting on the original training set. These issues with the labels then eventually affected the correctness scores of the data-to-text system's output \citep[an example of cascading of errors][]{castro_ferreira_etal_2019}. Leading to worse connections between the data input and text output, as is also found by the error analysis. Therefore, we believe our results are not in line with H3, nor H5 as the Dutch CACAPO corpus performed similarly to the other corpora, especially the English CACAPO corpus. These results are also not in line with previous research that found semi-supervised learning to especially bolster the results for small-scale datasets \citep{chang_etal_2021}. It might therefore be that the usage of language models nullifies the differences between datasets and their scale: previous work has shown that language models enables few-shot and zero-shot learning \citep{bender_etal_2021}, which could make the difference between dataset scale less important. The lack of difference between the Dutch and English datasets furthermore shows that, although the Dutch representation in mT5 is small (especially compared to the English representation in T5) it is still enough to overcome issues of undertraining.

\subsubsection*{Output Diversity}
The diversity differences between the various approaches were investigated using measures derived from \cite{vanmiltenburg_etal_2018}. We found support for H2: semi-supervised learning generally seemed to increase output diversity. For every investigated dataset, one of the semi-supervised learning approaches scored the highest on almost each of the diversity metrics. This is also in line with previous findings by \cite{kulhanek_etal_2021}. This suggests that semi-supervised learning can (at least partially) solve the previously highlighted ``catastrophic forgetting'' problem \citep{greco_etal_2019}, where a neural model is overfit too tightly during finetuning, which leads to the model forgetting about the diverse language in the language model.

The increase in diversity is relatively consistent amongst datasets (which does not support H4 that states that a bigger increase in diversity is expected for crowdsourced datasets) and dataset language (which does not support H5). However, the semi-supervised learning approach that leads to the largest diversity increases, differs per dataset. The pseudo-labeling approach is generally the most diversity-bolstering approach for CACAPO (nl) and WebNLG, whereas the data augmentation approach is the most effective approach for diversity increases for CACAPO (en) and E2E. However, it is difficult to pinpoint which characteristics of the datasets leads to these differences. Diversity differences may be due to the richness in content words in the original dataset (which leads to more perturbations with the data augmentation approach), or lack thereof. Alternatively, it can be that the extra texts used for the pseudo-labeling approach contain more (or less) diverse language for some datasets, leading to more (or less) diverse verbalizations of the data-to-text output. Future research is necessary to investigate which characteristics of the dataset---and extra data introduced by the semi-supervised learning approaches---are most salient when the goal is to increase text diversity.

\subsection{Future Work}
The current study found that semi-supervised learning is an effective technique for data-to-text generation, even when used in conjunction with a language model. The pseudo-labeling approach can increase both output diversity and quality, whereas the data augmentation approach is effective at increasing diversity, while keeping the quality consistent with a model only enriched with a language model. Furthermore, this result seems to be consistent amongst datasets. It may be that a good language model nullifies the differences between the original datasets, meaning that, for instance, the scale of the dataset does not matter as much. This in turn would mean that enriching training data with a language model can already be helpful for non-English NLG tasks, for which it is generally more difficult to find large-scale datasets \citep{riabi_etal_2021}. This would help to make neural NLG systems available for more people around the world without having to invest in large-scale datasets.

Two common semi-supervised learning approaches (pseudo-labeling and data augmentation) were explored in the design of the presented data-to-text system. However, for future research there are also other approaches to explore. For instance, data augmentation can be done in other ways, other than perturbing words with semantically similar words, as was done in this research. Also of interest would be \textit{entity resolution/text matching} (i.e., automatically connecting data with corresponding text), rather than pseudo-labeling \citep{ahmadi_etal_2021}. Many real-world companies have large collections of texts and related data, but are missing an explicit connection between the two \citep{vanderlee_etal_2020}. Such a task could help to produce large quantities of extra data, and may additionally help to tackle issues with mistakes in the original manual data annotations that could lead to bad texts. Especially interesting would be the use of unlabeled texts as an in-domain language model, rather than the Transformers-based language models such as BERT and T5 that were trained on an immense variety of different domains. \cite{sun_etal_2020} found that employing such an in-domain language model made the use of BERT unnecessary, and that the combination of in-domain pretraining and pseudo-labeling resulted in the best model for a text classification task. The impact of training big Transformers models such as BERT and T5 on CO$_2$ emissions is well documented \citep{strubell_etal_2019}. If the use of unlabeled texts for (generally smaller-scale) in-domain pretraining and pseudo-labeling is indeed also effective for NLG, this might help to focus development on more efficient and eco-friendly development of single-domain language models, rather than unsustainably striving for larger and larger all-encompassing Transformers models.

The findings of the current study highlight the feasibility of automatically extending datasets in a way that neural data-to-text systems benefit from the extension, even if the original dataset is small in size. This can be beneficial for robot journalism, for instance, that still mostly relies on rule-based NLG. Newspaper organisations oftentimes have large archives of texts and data, but aligning the two manually on a scale that permits a neural model to be trained on it, requires an amount of manpower that most newsrooms do not have available. By creating a small starting point and then using semi-supervised learning to create the full dataset, would be a less intensive undertaking---thus a smaller hurdle---for many companies to employ neural data-to-text systems.

\bibliographystyle{apacite}  
\bibliography{references}  

\newpage
\appendix
\section{Examples of output}
\begin{table*}[ht!]
\centering
\scriptsize
\begin{tabularx}{\textwidth}{ll|X|X}
\toprule
\bf Dataset & \bf Train type & \bf Input data & \bf Output text \\
\midrule
CACAPO (en) & No\_Ext & {\tiny victimAgeGroup="adult" \newline victimSex="female" \newline victimStatus="dead"} & Police say an adult female was found dead at the scene from apparent gunshot wounds. \\[25pt]
 & Dat\_Aug & {\tiny batterName="Xavier\_Nady" \newline pitchResult="scored" \newline teamName="Cubs" \newline inningNumber="sixth"} & Xavier Nady scored for the Cubs in the sixth \\[35pt]
 & Pseu\_Lab & {\tiny companyName="Microsoft" \newline stockChange="jumped" \newline stockChangePercentage="more\_than\_ 54\%" \newline moneyAmount1="\$138.40" \newline moneyAmount2="\$90.14" \newline timePoint="Jan.\_17"} & Microsoft shares jumped more than 54 \% to \$138.40, from \$90.14 on Jan. 17. \\
\midrule
CACAPO (nl) & No\_Ext & {\tiny incidentLocation="A12\_richting\_Arn-hem"\newline dateTime="dinsdag"\newline incidentType="dodelijk\_ongeluk"} & De A12 richting Arnhem is dinsdag door een dodelijk ongeluk geëindigd.\newline {\tiny \textit{The A12 to Arnhem has ended by a deadly accident on Tuesday.}} \\[35pt]
 & Dat\_Aug & {\tiny hasWonTeam="AZ" \newline matchDate="donderdagavond" \newline finalScore="1-2"  \newline hasLostTeam="ADO\_Den\_Haag"} & AZ heeft donderdagavond met 1-2 gewonnen van ADO Den Haag.\newline {\tiny \textit{AZ has won 1-2 against ADO Den Haag on Thursday night.}} \\[35pt]
 & Pseu\_Lab & {\tiny weatherType="regen" \newline compassDirection="noordoosten"} & De regen trekt verder naar het noordoosten. \newline {\tiny \textit{The rain will move further to the northeast.}} \\
\midrule
WebNLG & No\_Ext & {\tiny "Turkey" capital "Ankara" \newline "Turkey" leaderName "Ahmet Davutoğlu"} & The capital of Turkey is Ankara and the country's leader is Ahmet Davutoglu. \\[25pt]
 & Dat\_Aug & {\tiny Tim Brooke-Taylor" birthPlace "Derbyshire" \newline "Bananaman" starring "Tim Brooke-Taylor"} & Derbyshire born Tim Brooke Taylor starred in The film. \\[35pt]
 & Pseu\_Lab & {\tiny "Buzz Aldrin" occupation "Fighter pilot" \newline "Buzz Aldrin" was\_a\_crew\_member\_of "Apollo 11"} & He was a fighter pilot and a member of the Apollo 11 crew. \\
\midrule
E2E & No\_Ext & {\tiny name="The Punter" \newline eatType="pub" \newline food="English" \newline priceRange="high" \newline area="riverside" \newline familyFriendly="no" \newline near="Raja Indian Cuisine"} & The Punter is a high priced pub near Raja Indian Cuisine in the riverside area. It is not children friendly. \\[65pt]
 & Dat\_Aug & {\tiny name="Clowns" \newline eatType="pub" \newline customer rating="3 out of 5" \newline near="All Bar One"} & Clowns pub is a local - priced, local - rated, and - friendly - friendly coffee bar located in the city market, near All Bar One. \\[35pt]
 & Pseu\_Lab & {\tiny name="The Cricketers" \newline eatType="restaurant" \newline customer rating="low" \newline familyFriendly="no" \newline near="Ranch"} & The Cricketers is a restaurant located near Rancho Mexican Cafe. It is not family-friendly. \\
\bottomrule
\end{tabularx}
\renewcommand{\thetable}{A.1}
\caption{Random examples of output for each of the datasets and semi-supervised learning methods.}
\label{tab:exampleoutputs}
\end{table*}

\newpage
\section{Evaluation results per dataset domain}
\begin{table*}[ht!]
\centering
\renewcommand{\arraystretch}{0.86}
\scriptsize
\begin{tabular*}{\textwidth}{l @{\extracolsep{\fill}} llrrrrr}
\toprule
\textbf{Dataset} & \textbf{Domain} & \textbf{Train type} & \textbf{BLEU} & \textbf{NIST} & \textbf{BertScore} & \textbf{METEOR} & \textbf{ROUGE-L} \\
\midrule
\multirow{3}{*}{CACAPO}  & \multirow{3}{*}{\texttt{Incidents} (en)}  & No\_Ext     & 29.47 & 5.11  & 58.55     & 51.28  & 47.08   \\
 &  & Dat\_Aug    & 28.15 & 4.99  & 53.71     & 47.18  & 42.92   \\
 &  & Pseu\_Lab    & \textbf{34.51} & \textbf{5.52}  & \textbf{62.06} & \textbf{54.75} & \textbf{52.07} \\
\midrule
\multirow{3}{*}{CACAPO}  & \multirow{3}{*}{\texttt{Sports} (en)}     & No\_Ext     & 31.77 & 6.63  & 61.00     & 56.43  & 52.73   \\
 &  & Dat\_Aug    & 23.30 & 5.80  & 52.57     & 48.76  & 46.35   \\
 &  & Pseu\_Lab    & \textbf{37.28} & \textbf{7.23}  & \textbf{65.55} & \textbf{60.02} & \textbf{57.30} \\
\midrule
\multirow{3}{*}{CACAPO}  & \multirow{3}{*}{\texttt{Stocks} (en)}     & No\_Ext     & 25.95 & 4.94  & 50.03     & 50.98  & 45.62   \\
 &  & Dat\_Aug    & 23.02 & 5.14  & 44.50     & 48.23  & 42.55   \\
 &  & Pseu\_Lab    & \textbf{32.44} & \textbf{6.05}  & \textbf{54.24} & \textbf{55.08} & \textbf{50.72} \\
\midrule
\multirow{3}{*}{CACAPO}  & \multirow{3}{*}{\texttt{Weather} (en)}    & No\_Ext     & 33.95 & 6.24  & 65.88     & 62.61  & 56.36   \\
 &  & Dat\_Aug    & 25.80 & 5.29  & 57.27     & 50.29  & 49.71   \\
 &  & Pseu\_Lab    & \textbf{39.59} & \textbf{6.70}  & \textbf{70.34} & \textbf{66.99} & \textbf{62.97} \\
\midrule
\multirow{3}{*}{CACAPO} & \multirow{3}{*}{\texttt{Incidents} (nl)}  & No\_Ext     & 34.22 & 5.39  & 85.15     & 53.38  & 51.53   \\
 &  & Dat\_Aug    & 36.71 & 5.01  & 87.20     & 55.50  & 57.10   \\
 &  & Pseu\_Lab    & \textbf{43.39} & \textbf{6.54} & \textbf{87.65} & \textbf{60.20} & \textbf{60.10} \\
\midrule
\multirow{3}{*}{CACAPO} & \multirow{3}{*}{\texttt{Sports} (nl)}     & No\_Ext     & 18.05 & 4.40  & 81.75     & 41.35  & 42.27   \\
 &  & Dat\_Aug    & 18.45 & 4.35  & 81.75     & 42.13  & 43.04   \\
 &  & Pseu\_Lab    & \textbf{25.45} & \textbf{5.19} & \textbf{83.42} & \textbf{48.11} & \textbf{48.32} \\
\midrule
\multirow{3}{*}{CACAPO} & \multirow{3}{*}{\texttt{Stocks} (nl)}     & No\_Ext     & 51.69 & 7.84  & 88.51     & 67.00  & 64.75   \\
 &  & Dat\_Aug    & 45.04 & 7.40  & 87.63     & 64.91  & 62.33   \\
 &  & Pseu\_Lab    & \textbf{63.31} & \textbf{8.98} & \textbf{91.03} & \textbf{74.83} & \textbf{72.75} \\
\midrule
\multirow{3}{*}{CACAPO} & \multirow{3}{*}{\texttt{Weather} (nl)}    & No\_Ext     & 24.28 & 4.61  & 82.50     & 47.29  & 46.44   \\
 &  & Dat\_Aug    & 49.07 & 7.32  & 90.42     & 69.39  & 67.75   \\
 &  & Pseu\_Lab    & \textbf{76.66} & \textbf{10.07} & \textbf{95.37} & \textbf{84.84} & \textbf{84.36} \\
\midrule
\multirow{3}{*}{WebNLG}  & \multirow{3}{*}{\texttt{Airport}}         & No\_Ext     & \textbf{51.54} & \textbf{7.35} & \textbf{74.63} & \textbf{74.80} & \textbf{61.17} \\
 & & Dat\_Aug    & 25.94 & 4.66  & 48.51     & 50.72  & 42.62   \\
 & & Pseu\_Lab    & 50.61 & 7.16  & 72.29     & 72.75  & 60.46   \\
\midrule
\multirow{3}{*}{WebNLG}  & \multirow{3}{*}{\texttt{Astronaut}}       & No\_Ext     & \textbf{48.89} & \textbf{6.54} & \textbf{77.98} & \textbf{76.03} & \textbf{65.04} \\
 &  & Dat\_Aug    & 21.04 & 3.82  & 55.69     & 47.77  & 43.08   \\
 &  & Pseu\_Lab    & 45.48 & 6.27  & 73.41     & 72.45  & 60.98   \\
\midrule
\multirow{3}{*}{WebNLG} & \multirow{3}{*}{\texttt{Building}}        & No\_Ext     & \textbf{53.48} & \textbf{7.84} & \textbf{76.93} & \textbf{78.06} & \textbf{65.97} \\
 &  & Dat\_Aug    & 34.51 & 5.77  & 60.05     & 62.70  & 52.65   \\
 &  & Pseu\_Lab    & 50.46 & 7.59  & 73.84     & 74.87  & 63.64   \\
\midrule
\multirow{3}{*}{WebNLG}  & \multirow{3}{*}{\texttt{City}}            & No\_Ext     & \textbf{29.01} & 2.50  & \textbf{52.93}     & 53.23  & \textbf{47.14}   \\
 & & Dat\_Aug    & 21.09 & 2.42  & 47.79     & 47.46  & 41.03   \\
 &  & Pseu\_Lab    & 27.83 & \textbf{4.18}  & 52.00     & \textbf{54.64}  & 44.84   \\
\midrule
\multirow{3}{*}{WebNLG}  & \multirow{3}{*}{\texttt{ComicsCharacter}} & No\_Ext     & \textbf{48.95} & \textbf{6.31}  & \textbf{72.50}     & 76.11  & 58.69   \\
 & & Dat\_Aug    & 38.06 & 5.56  & 61.60     & 69.43  & 51.17   \\
 & & Pseu\_Lab    & 48.74 & 6.22  & 70.92     & \textbf{76.68}  & \textbf{62.42}   \\
\midrule
\multirow{3}{*}{WebNLG} & \multirow{3}{*}{\texttt{Food}}            & No\_Ext     & \textbf{46.81} & \textbf{7.30}  & \textbf{71.35}     & \textbf{70.53}  & \textbf{56.11}   \\
 &  & Dat\_Aug    & 23.84 & 4.57  & 46.37     & 47.74  & 40.19   \\
 & & Pseu\_Lab    & 38.10 & 6.28  & 62.91     & 61.87  & 48.89   \\
\midrule
\multirow{3}{*}{WebNLG} & \multirow{3}{*}{\texttt{Monument}}        & No\_Ext     & 43.97 & 5.95  & 69.70     & 71.89  & 56.88   \\
 &  & Dat\_Aug    & 27.88 & 4.39  & 53.21     & 54.12  & 46.25   \\
 & & Pseu\_Lab    & \textbf{45.38} & \textbf{6.10}  & \textbf{71.26}     & \textbf{73.17}  & \textbf{58.58}   \\
\midrule
\multirow{3}{*}{WebNLG} & \multirow{3}{*}{\texttt{SportsTeam}}      & No\_Ext     & \textbf{46.07} & \textbf{6.89}  & \textbf{72.41}     & \textbf{72.59}  & \textbf{59.58}   \\
 & & Dat\_Aug    & 25.89 & 4.79  & 52.68     & 52.20  & 42.65   \\
 & & Pseu\_Lab    & 43.88 & 6.77  & 69.53     & 70.94  & 58.18   \\
\midrule
\multirow{3}{*}{WebNLG} & \multirow{3}{*}{\texttt{University}}     & No\_Ext     & \textbf{60.07} & \textbf{7.34}  & \textbf{78.10}     & \textbf{78.12}  & \textbf{68.93}   \\
 & & Dat\_Aug    & 29.60 & 4.59  & 52.81     & 53.72  & 48.12   \\
 & & Pseu\_Lab    & 55.34 & 6.89  & 74.66     & 75.65  & 64.27   \\
\midrule
\multirow{3}{*}{WebNLG} & \multirow{3}{*}{\texttt{WrittenWork}}     & No\_Ext     & \textbf{54.39} & \textbf{7.45}  & \textbf{75.86}     & \textbf{77.09}  & \textbf{65.23}   \\
 & & Dat\_Aug    & 36.00 & 5.60  & 62.28     & 61.89  & 53.28   \\
 & & Pseu\_Lab    & 52.25 & 7.16  & 73.16     & 74.57  & 62.48   \\
\midrule
\multirow{3}{*}{E2E}     &                 & No\_Ext     & \textbf{66.05} & \textbf{7.08}  & \textbf{79.40}     & \textbf{80.21}  & \textbf{44.97}   \\
 &                 & Dat\_Aug    & 28.41 & 4.15  & 56.41     & 62.49  & 33.40   \\
 &                 & Pseu\_Lab    & 50.51 & 4.65  & 63.12     & 60.39  & 38.48  \\
\bottomrule
\end{tabular*}
\renewcommand{\thetable}{B.1}
\caption{Automatic metric results of the different (XL-format) semi-supervised learning approaches (No\_Ext = no training set extension, Dat\_Aug = data augmentation, Pseu\_Lab = pseudo-labeling) for each dataset domain (bold = highest).}
\label{tab:autmetsslm}
\end{table*}

\begin{table*}[ht!]
\centering
\setlength{\tabcolsep}{4.5pt}
\renewcommand{\arraystretch}{0.95}
\scriptsize
\begin{tabular}{lllrrrrrrrrr}
\toprule
\textbf{Dataset} & \textbf{Domain}        & \textbf{Train type} & \textbf{ASL}   & \textbf{SDSL}  & \textbf{Types} & \textbf{TTR\textsubscript{1}} & \textbf{TTR\textsubscript{2}} & \textbf{\%Novel} & \textbf{Cov}  & \textbf{Nov}  & \textbf{Loc\textsubscript{1}} \\
\midrule
\multirow{3}{*}{CACAPO}  & \multirow{3}{*}{\texttt{Incidents} (en)}  & No\_Ext     & 16.05 & 6.76  & 656   & 0.63 & 0.90 & 99.04  & 0.53 & 0.19 & 0.49 \\
 & & Dat\_Aug    & \textbf{17.83} & 7.46  & \textbf{844}   & \textbf{0.68} & \textbf{0.94} & \textbf{100.00} & \textbf{0.62} & \textbf{0.31} & 0.47 \\
 & & Pseu\_Lab   & 15.86 & \textbf{8.47}  & 696   & 0.63 & 0.89 & 99.04  & 0.57 & 0.20 & \textbf{0.52} \\
\midrule
\multirow{3}{*}{CACAPO}  & \multirow{3}{*}{\texttt{Sports} (en)}     & No\_Ext     & 19.32 & 7.95  & 1499  & 0.69 & \textbf{0.96} & 99.84  & 0.58 & 0.14 & 0.57 \\
 & & Dat\_Aug    & 19.07 & 8.17  & \textbf{1574}  & \textbf{0.70} & \textbf{0.96} & \textbf{100.00} & 0.59 & \textbf{0.16} & 0.54 \\
 & & Pseu\_Lab   & \textbf{19.40} & \textbf{8.52}  & 1559  & \textbf{0.70} & \textbf{0.96} & \textbf{100.00} & \textbf{0.61} & 0.13 & \textbf{0.61} \\
\midrule
\multirow{3}{*}{CACAPO}  & \multirow{3}{*}{\texttt{Stocks} (en)} & No\_Ext     & 18.15 & 9.51  & 1369  & 0.65 & 0.91 & 99.07  & 0.49 & 0.31 & 0.43 \\
 & & Dat\_Aug    & 19.10 & 9.32  & \textbf{1581}  & \textbf{0.68} & \textbf{0.94} & \textbf{100.00} & \textbf{0.54} & \textbf{0.38} & 0.45 \\
 & & Pseu\_Lab   & \textbf{19.76} & \textbf{11.10} & 1556  & 0.65 & 0.91 & 99.69  & 0.53 & \textbf{0.38} & \textbf{0.48} \\
\midrule
\multirow{3}{*}{CACAPO}  & \multirow{3}{*}{\texttt{Weather} (en)}    & No\_Ext     & \textbf{13.62} & 7.05  & 746   & 0.62 & 0.91 & 94.28  & 0.57 & 0.14 & 0.59 \\
 & & Dat\_Aug    & 12.75 & \textbf{7.22}  & 757   & \textbf{0.65} & \textbf{0.92} & \textbf{99.18}  & 0.57 & \textbf{0.16} & 0.57 \\
 & & Pseu\_Lab   & 13.40 & 7.10  & \textbf{768}   & 0.63 & 0.91 & 94.55  & \textbf{0.60} & 0.14 & \textbf{0.63} \\
\midrule
\multirow{3}{*}{CACAPO}  & \multirow{3}{*}{\texttt{Incidents} (nl)}  & No\_Ext     & 13.84 & 5.24  & 539   & 0.58 & 0.86 & \textbf{100.00} & 0.53 & 0.21 & 0.52 \\
 & & Dat\_Aug    & 12.34 & 4.94  & 526   & 0.58 & \textbf{0.87} & \textbf{100.00} & 0.51 & 0.22 & 0.54 \\
 & & Pseu\_Lab   & \textbf{14.60} & \textbf{6.51}  & \textbf{584}   & \textbf{0.59} & \textbf{0.87} & 99.01  & \textbf{0.58} & \textbf{0.23} & \textbf{0.58} \\
\midrule
\multirow{3}{*}{CACAPO}  & \multirow{3}{*}{\texttt{Sports} (nl)}     & No\_Ext     & 13.78 & \textbf{5.82}  & 920   & 0.66 & 0.93 & \textbf{100.00} & 0.48 & 0.10 & 0.42 \\
 & & Dat\_Aug    & 13.43 & 5.50  & 990   & 0.67 & 0.94 & \textbf{100.00} & 0.49 & 0.13 & 0.43 \\
 & & Pseu\_Lab   & \textbf{13.90} & 5.69  & \textbf{1127}  & \textbf{0.70} & \textbf{0.96} & 99.72  & \textbf{0.56} & \textbf{0.14} & \textbf{0.48} \\
\midrule
\multirow{3}{*}{CACAPO}  & \multirow{3}{*}{\texttt{Stocks} (nl)}     & No\_Ext     & 15.35 & 7.18  & 1373  & 0.65 & 0.91 & 97.54  & 0.58 & 0.31 & 0.62 \\
 & & Dat\_Aug    & 15.17 & 7.22  & 1325  & 0.65 & 0.92 & \textbf{99.11}  & 0.54 & 0.32 & 0.60 \\
 & & Pseu\_Lab   & \textbf{15.78} & \textbf{7.26}  & \textbf{1584}  & \textbf{0.67} & \textbf{0.93} & 96.20  & \textbf{0.65} & \textbf{0.37} & \textbf{0.70} \\
\midrule
\multirow{3}{*}{CACAPO}  & \multirow{3}{*}{\texttt{Weather} (nl)} & No\_Ext     & 15.07 & \textbf{10.27} & 289   & 0.42 & 0.70 & \textbf{98.75}  & 0.52 & 0.06 & 0.51 \\
 & & Dat\_Aug    & 15.14 & 5.35  & \textbf{459}   & 0.59 & \textbf{0.88} & 97.24  & 0.76 & \textbf{0.17} & 0.73 \\
 & & Pseu\_Lab   & \textbf{15.29} & 5.31  & 446   & \textbf{0.60} & \textbf{0.88} & 82.21  & \textbf{0.84} & 0.06 & \textbf{0.86} \\
\midrule
\multirow{3}{*}{WebNLG}  & \multirow{3}{*}{\texttt{Airport}} & No\_Ext     & 16.77 & 6.21  & 425   & 0.43 & 0.70 & 78.17  & 0.65 & 0.03 & \textbf{0.74} \\
 & & Dat\_Aug    & \textbf{16.89} & \textbf{8.23}  & 417   & 0.42 & 0.70 & \textbf{99.30}  & 0.50 & \textbf{0.17} & 0.48 \\
 & & Pseu\_Lab   & 16.78 & 6.19  & \textbf{474}   & \textbf{0.45} & \textbf{0.71} & 79.23  & \textbf{0.68} & 0.08 & 0.71 \\
\midrule
\multirow{3}{*}{WebNLG}  & \multirow{3}{*}{\texttt{Astronaut}} & No\_Ext     & 17.56 & 7.42  & 208   & 0.47 & 0.69 & 74.68  & 0.53 & 0.02 & \textbf{0.71} \\
 & & Dat\_Aug    & 16.66 & 7.26  & 231   & 0.45 & 0.69 & \textbf{98.70}  & 0.40 & \textbf{0.21} & 0.36 \\
 & & Pseu\_Lab   & \textbf{17.79} & \textbf{7.65}  & \textbf{259}   & \textbf{0.51} & \textbf{0.77} & 76.62  & \textbf{0.57} & 0.12 & 0.68 \\
\midrule
\multirow{3}{*}{WebNLG}  & \multirow{3}{*}{\texttt{Building}} & No\_Ext     & 17.60 & 6.61  & 458   & 0.44 & 0.70 & 78.26  & \textbf{0.72} & 0.02 & \textbf{0.79} \\
 & & Dat\_Aug    & \textbf{17.72} & \textbf{6.93}  & 456   & 0.43 & 0.69 & \textbf{98.81}  & 0.63 & \textbf{0.12} & 0.60 \\
 & & Pseu\_Lab   & 17.29 & 6.52  & \textbf{473}   & \textbf{0.46} & \textbf{0.73} & 76.68  & \textbf{0.72} & 0.05 & 0.76 \\
\midrule
\multirow{3}{*}{WebNLG}  & \multirow{3}{*}{\texttt{City}} & No\_Ext     & 11.24 & 2.79  & 182   & 0.32 & 0.52 & 79.35  & 0.69 & 0.11 & 0.51 \\
 & & Dat\_Aug    & 11.69 & 3.14  & 208   & 0.37 & 0.59 & \textbf{92.90}  & 0.69 & 0.23 & 0.47 \\
 & & Pseu\_Lab   & \textbf{13.79} & \textbf{4.66}  & \textbf{270}   & \textbf{0.38} & \textbf{0.64} & 87.10  & \textbf{0.81} & \textbf{0.37} & \textbf{0.55} \\
\midrule
\multirow{3}{*}{WebNLG}  & \multirow{3}{*}{\texttt{ComicsChar}} & No\_Ext     & 14.97 & \textbf{5.69}  & 174   & 0.45 & 0.73 & 95.16  & 0.64 & 0.03 & \textbf{0.70} \\
 & & Dat\_Aug    & 14.97 & 5.02  & 172   & 0.45 & 0.71 & \textbf{100.00} & 0.60 & \textbf{0.06} & 0.60 \\
 & & Pseu\_Lab   & \textbf{15.39} & 5.50  & \textbf{188}   & \textbf{0.49} & \textbf{0.78} & 95.16  & \textbf{0.66} & \textbf{0.06} & 0.68 \\
\midrule
\multirow{3}{*}{WebNLG}  & \multirow{3}{*}{\texttt{Food}} & No\_Ext     & 15.74 & \textbf{7.53}  & 439   & 0.42 & 0.71 & 84.10  & 0.67 & 0.01 & \textbf{0.68} \\
 & & Dat\_Aug    & \textbf{15.91} & 7.36  & \textbf{585}   & \textbf{0.45} & \textbf{0.78} & \textbf{98.97}  & 0.66 & \textbf{0.24} & 0.41 \\
 & & Pseu\_Lab   & 15.61 & 7.22  & 517   & 0.44 & 0.72 & 88.97  & \textbf{0.71} & 0.09 & 0.59 \\
\midrule
\multirow{3}{*}{WebNLG}  & \multirow{3}{*}{\texttt{Monument}} & No\_Ext     & 19.05 & 7.35  & 145   & \textbf{0.44} & 0.71 & 73.86  & 0.62 & 0.04 & 0.74 \\
 & & Dat\_Aug    & 18.49 & \textbf{7.74}  & \textbf{176}   & \textbf{0.44} & \textbf{0.72} & \textbf{98.86}  & 0.55 & \textbf{0.26} & 0.51 \\
 & & Pseu\_Lab   & \textbf{19.58} & 7.59  & 152   & 0.42 & 0.69 & 72.73  & \textbf{0.65} & 0.04 & \textbf{0.75} \\
\midrule
\multirow{3}{*}{WebNLG}  & \multirow{3}{*}{\texttt{SportsTeam}} & No\_Ext     & \textbf{15.60} & 5.46  & 352   & 0.47 & 0.72 & 88.56  & 0.64 & 0.02 & \textbf{0.66} \\
 & & Dat\_Aug    & 15.46 & \textbf{5.69}  & 362   & 0.44 & 0.71 & \textbf{100.00} & 0.57 & \textbf{0.11} & 0.46 \\
 & & Pseu\_Lab   & 15.31 & 5.53  & \textbf{383}   & \textbf{0.48} & \textbf{0.75} & 82.59  & \textbf{0.68} & 0.04 & 0.63 \\
\midrule
\multirow{3}{*}{WebNLG}  & \multirow{3}{*}{\texttt{University}} & No\_Ext     & 16.47 & 6.78  & 188   & 0.43 & 0.69 & 67.35  & 0.60 & 0.02 & \textbf{0.75} \\
 & & Dat\_Aug    & 14.86 & 6.53  & 227   & 0.46 & 0.76 & \textbf{95.24}  & 0.53 & \textbf{0.22} & 0.47 \\
 & & Pseu\_Lab   & \textbf{17.30} & \textbf{7.32}  & \textbf{255}   & \textbf{0.48} & \textbf{0.77} & 65.31  & \textbf{0.73} & 0.12 & 0.72 \\
\midrule
\multirow{3}{*}{WebNLG}  & \multirow{3}{*}{\texttt{Writ.Work}} & No\_Ext     & 18.15 & 6.78  & 352   & 0.41 & 0.65 & 84.21  & 0.61 & 0.01 & \textbf{0.74} \\
 & & Dat\_Aug    & 17.74 & 6.49  & 397   & 0.44 & \textbf{0.71} & \textbf{97.98}  & 0.58 & \textbf{0.12} & 0.59 \\
 & & Pseu\_Lab   & \textbf{18.32} & \textbf{6.97}  & \textbf{400}   & \textbf{0.45} & \textbf{0.71} & 82.59  & \textbf{0.67} & 0.04 & 0.71 \\
\midrule
\multirow{3}{*}{E2E}     &                 & No\_Ext     & 28.58 & \textbf{7.66}  & 120   & 0.34 & 0.50 & \textbf{100.00} & 0.11 & 0.00 & \textbf{0.11} \\
 &                 & Dat\_Aug    & \textbf{34.42} & 7.73  & \textbf{223}   & \textbf{0.38} & \textbf{0.55} & \textbf{100.00} & \textbf{0.16} & \textbf{0.03} & 0.10 \\
 &                 & Pseu\_Lab   & 23.22 & 5.26  & 115   & 0.26 & 0.38 & \textbf{100.00} & 0.07 & \textbf{0.03} & 0.08 \\
\bottomrule
\end{tabular}
\renewcommand{\thetable}{B.2}
\caption{Average sentence length, standard deviation of sentence length, mean-segmented type-token ratio (TTR), bigram TTR, percentage novel descriptions, coverage, novelty and local recall with importance class 1 (bold = highest).}
\label{tab:divmetsslm}
\end{table*}

\begin{table*}
\centering
\renewcommand{\arraystretch}{0.87}
\scriptsize
\begin{tabular*}{\textwidth}{l @{\extracolsep{\fill}} llrrrr}
\toprule
\bf Dataset & \bf Domain        & \bf Train Type & \bf N  & \bf Fluency     & \bf Correctness & \bf Grammaticality \\
\midrule
\multirow{3}{*}{CACAPO} & \multirow{3}{*}{\texttt{Incidents} (en)} & No\_Ext    & \multirow{3}{*}{11} & 4.90 (2.16)$^a$ & 4.77 (1.97)$^a$ & 2.93 (0.95)$^a$    \\
 & & Dat\_Aug   & & 4.77 (2.18)$^a$ & 4.77 (2.00)$^a$ & 2.86 (1.00)$^a$    \\
 & & Pseu\_Lab  & & \textbf{5.99 (1.57)$^b$} & \textbf{5.75 (1.82)$^b$} & \textbf{3.41 (0.77)$^b$}    \\
\midrule
\multirow{3}{*}{CACAPO}  & \multirow{3}{*}{\texttt{Sports} (en)} & No\_Ext    & \multirow{3}{*}{12} & 4.51 (2.03)$^a$ & 4.56 (1.80)$^a$ & 2.83 (0.91)$^a$    \\
 & & Dat\_Aug   & & 4.59 (2.05)$^a$ & 4.60 (1.92)$^{a,b}$ & 2.91 (0.91)$^a$    \\
 & & Pseu\_Lab  & & \textbf{5.42 (1.88)$^b$} & \textbf{5.23 (1.82)$^b$} & \textbf{3.24 (0.84)$^b$}    \\
\midrule
\multirow{3}{*}{CACAPO} & \multirow{3}{*}{\texttt{Stocks} (en)} & No\_Ext    & \multirow{3}{*}{10} & 4.95 (2.09)$^a$ & 5.23 (1.79)$^a$ & 3.06 (1.07)$^a$    \\
 & & Dat\_Aug   & & 4.80 (2.10)$^a$ & 5.12 (1.79)$^a$ & 3.02 (1.02)$^a$    \\
 & & Pseu\_Lab  & & \textbf{5.80 (1.58)$^b$} & \textbf{5.82 (1.46)$^a$} & \textbf{3.42 (0.78)$^b$}    \\
\midrule
\multirow{3}{*}{CACAPO} & \multirow{3}{*}{\texttt{Weather} (en)} & No\_Ext    & \multirow{3}{*}{10} & 5.02 (1.85)$^a$ & 5.24 (1.78)$^a$ & 2.81 (0.99)$^a$    \\
 & & Dat\_Aug   & & 5.17 (1.81)$^a$ & 5.29 (1.76)$^a$ & 2.94 (0.98)$^a$    \\
 & & Pseu\_Lab  & & \textbf{5.48 (1.67)$^a$} & \textbf{5.54 (1.57)$^a$} & \textbf{3.10 (0.99)$^a$}    \\
\midrule
\multirow{3}{*}{CACAPO} & \multirow{3}{*}{\texttt{Incidents} (nl)} & No\_Ext    & \multirow{3}{*}{11} & 4.87 (2.05)$^a$ & 5.23 (1.85)$^a$ & 2.84 (1.10)$^a$    \\
 & & Dat\_Aug   & & 4.87 (2.14)$^a$ & 5.25 (1.86)$^a$ & 2.84 (1.11)$^a$    \\
 & & Pseu\_Lab  & & \textbf{6.29 (1.20)$^b$} & \textbf{5.91 (1.40)$^a$} & \textbf{3.61 (0.66)$^b$}    \\
\midrule
\multirow{3}{*}{CACAPO} & \multirow{3}{*}{\texttt{Sports} (nl)} & No\_Ext    & \multirow{3}{*}{10} & 4.73 (2.12)$^a$ & 5.03 (1.98)$^a$ & 2.90 (1.10)$^a$    \\
 & & Dat\_Aug   & & 4.75 (2.08)$^a$ & 4.98 (1.92)$^a$ & 2.90 (1.10)$^a$    \\
 & & Pseu\_Lab  & & \textbf{5.44 (1.75)$^a$} & \textbf{5.34 (1.77)$^a$} & \textbf{3.22 (0.96)$^a$}    \\
\midrule
\multirow{3}{*}{CACAPO} & \multirow{3}{*}{\texttt{Stocks} (nl)} & No\_Ext    & \multirow{3}{*}{11} & 5.00 (1.98)$^a$ & 5.32 (1.77)$^a$ & 3.08 (0.97)$^a$    \\
 & & Dat\_Aug   & & 4.95 (2.04)$^a$ & 5.28 (1.82)$^a$ & 2.95 (1.08)$^a$    \\
 & & Pseu\_Lab  & & \textbf{5.91 (1.57)$^b$} & \textbf{5.67 (1.64)$^a$} & \textbf{3.59 (0.69)$^b$}    \\
\midrule
\multirow{3}{*}{CACAPO} & \multirow{3}{*}{\texttt{Weather} (nl)} & No\_Ext    & \multirow{3}{*}{9} & 4.31 (1.90)$^a$ & 4.97 (2.03)$^a$ & 2.45 (1.10)$^a$    \\
 & & Dat\_Aug   & & 4.27 (2.02)$^a$ & 5.06 (1.92)$^a$ & 2.38 (1.13)$^a$    \\
 & & Pseu\_Lab  & & \textbf{6.10 (1.41)$^b$} & \textbf{6.12 (1.40)$^b$} & \textbf{3.52 (0.75)$^b$}    \\
\midrule
\multirow{3}{*}{WebNLG} & \multirow{3}{*}{\texttt{Airport}} & No\_Ext    & \multirow{3}{*}{11} & 4.23 (2.06)$^a$ & 4.00 (2.12)$^a$ & 2.85 (1.02)$^a$    \\
 & & Dat\_Aug   & & 4.23 (2.08)$^a$ & 3.99 (2.19)$^a$ & 2.84 (1.00)$^a$    \\
 & & Pseu\_Lab  & & \textbf{5.44 (1.71)$^b$} & \textbf{5.40 (1.82)$^b$} & \textbf{3.39 (0.83)$^b$}    \\
\midrule
\multirow{3}{*}{WebNLG} & \multirow{3}{*}{\texttt{Astronaut}} & No\_Ext    & \multirow{3}{*}{12} & 4.93 (2.01)$^a$ & 3.13 (1.86)$^a$ & 3.06 (0.96)$^a$    \\
 & & Dat\_Aug   & & 4.89 (1.98)$^a$ & 3.02 (1.85)$^a$ & 2.99 (1.00)$^a$    \\
 & & Pseu\_Lab  & & \textbf{6.26 (1.17)$^b$} & \textbf{6.17 (1.37)$^b$} & \textbf{3.70 (0.58)$^b$}    \\
\midrule
\multirow{3}{*}{WebNLG} & \multirow{3}{*}{\texttt{Building}} & No\_Ext    & \multirow{3}{*}{6} & 5.23 (1.57)$^a$ & 4.84 (1.72)$^a$ & 3.07 (0.82)$^a$    \\
 & & Dat\_Aug   & & 5.26 (1.61)$^a$ & 4.79 (1.83)$^a$ & 3.08 (0.76)$^a$    \\
 & & Pseu\_Lab  & & \textbf{6.03 (1.34)$^a$} & \textbf{6.14 (1.33)$^b$} & \textbf{3.52 (0.69)$^a$}    \\
\midrule
\multirow{3}{*}{WebNLG} & \multirow{3}{*}{\texttt{City}} & No\_Ext    & \multirow{3}{*}{9} & \textbf{5.42 (1.88)$^a$} & 4.81 (2.08)$^a$ & \textbf{3.27 (1.00)$^a$}    \\
 & & Dat\_Aug   & & 5.40 (1.95)$^a$ & \textbf{4.84 (2.02)$^a$} & \textbf{3.27 (0.99)$^a$}    \\
 & & Pseu\_Lab  & & 5.18 (2.07)$^a$ & 4.44 (2.27)$^a$ & 3.20 (0.96)$^a$    \\
\midrule
\multirow{3}{*}{WebNLG} & \multirow{3}{*}{\texttt{ComicsChar}} & No\_Ext    & \multirow{3}{*}{12} & 5.66 (1.75)$^a$ & 5.46 (1.78)$^a$ & 3.37 (0.81)$^a$    \\
 & & Dat\_Aug   & & \textbf{5.73 (1.70)$^a$} & 5.48 (1.80)$^a$ & \textbf{3.39 (0.83)$^a$}    \\
 & & Pseu\_Lab  & & 5.72 (1.76)$^a$ & \textbf{5.59 (1.84)$^a$} & 3.38 (0.84)$^a$    \\
\midrule
\multirow{3}{*}{WebNLG} & \multirow{3}{*}{\texttt{Food}} & No\_Ext    & \multirow{3}{*}{8} & 3.35 (2.28)$^a$ & 2.92 (2.18)$^a$ & 2.15 (1.13)$^a$    \\
 & & Dat\_Aug   & & 3.35 (2.29)$^a$ & 2.97 (2.09)$^a$ & 2.14 (1.14)$^a$    \\
 & & Pseu\_Lab  & & \textbf{5.89 (1.69)$^b$} & \textbf{5.46 (1.96)$^b$} & \textbf{3.41 (0.91)$^b$}    \\
\midrule
\multirow{3}{*}{WebNLG} & \multirow{3}{*}{\texttt{Monument}} & No\_Ext    & \multirow{3}{*}{12} & 3.99 (2.13)$^a$ & 2.80 (2.10)$^a$ & 2.50 (0.97)$^a$    \\
 & & Dat\_Aug   & & 4.13 (2.17)$^a$ & 2.99 (2.19)$^a$ & 2.51 (1.02)$^a$    \\
 & & Pseu\_Lab  & & \textbf{5.92 (1.44)$^b$} & \textbf{5.83 (1.75)$^b$} & \textbf{3.42 (0.72)$^b$}    \\
\midrule
\multirow{3}{*}{WebNLG} & \multirow{3}{*}{\texttt{SportsTeam}} & No\_Ext    & \multirow{3}{*}{11} & 4.89 (1.89)$^a$ & 4.64 (2.16)$^a$ & 3.06 (0.97)$^a$    \\
 & & Dat\_Aug   & & 4.90 (1.87)$^a$ & 4.66 (2.16)$^a$ & 3.04 (0.92)$^a$    \\
 & & Pseu\_Lab  & & \textbf{5.96 (1.38)$^b$} & \textbf{5.94 (1.54)$^b$} & \textbf{3.51 (0.69)$^b$}    \\
\midrule
\multirow{3}{*}{WebNLG} & \multirow{3}{*}{\texttt{University}} & No\_Ext    & \multirow{3}{*}{9} & 4.60 (1.94)$^a$ & 3.71 (2.03)$^a$ & 3.09 (0.85)$^a$    \\
 & & Dat\_Aug   & & 4.70 (1.82)$^a$ & 3.60 (2.02)$^a$ & 3.07 (0.79)$^a$    \\
 & & Pseu\_Lab  & & \textbf{5.70 (1.72)$^b$} & \textbf{5.60 (1.84)$^b$} & \textbf{3.49 (0.73)$^b$}    \\
\midrule
\multirow{3}{*}{WebNLG} & \multirow{3}{*}{\texttt{Writ.Work}} & No\_Ext    & \multirow{3}{*}{9} & 4.00 (2.05)$^a$ & 3.82 (2.16)$^a$ & 2.68 (1.02)$^a$    \\
 & & Dat\_Aug   & & 4.00 (2.07)$^a$ & 3.91 (2.22)$^a$ & 2.71 (1.02)$^a$    \\
 & & Pseu\_Lab  & & \textbf{5.21 (1.92)$^b$} & \textbf{5.66 (1.78)$^b$} & \textbf{3.23 (0.88)$^b$}    \\
\midrule
\multirow{3}{*}{E2E}     & & No\_Ext    & \multirow{3}{*}{9} & 5.08 (1.73)$^a$ & \textbf{4.25 (1.86)$^a$} & 3.06 (0.88)$^a$    \\
 & & Dat\_Aug   & & 5.06 (1.70)$^a$ & 4.20 (1.83)$^a$ & 2.98 (0.91)$^a$    \\
 & & Pseu\_Lab  & & \textbf{6.26 (1.02)$^b$} & 3.86 (1.71)$^a$ & \textbf{3.61 (0.65)$^b$}   \\
\bottomrule
\end{tabular*}
\renewcommand{\thetable}{B.3}
\caption{Mean fluency, correctness, and grammaticality per semi-supervised learning type for each domain (SDs between brackets). Different superscripts indicate significant differences between semi-supervised learning methods for that domain. Higher scores mean more positively perceived output.}
\label{tab:meanhumevalperdomain}
\end{table*}
\clearpage
\section{Multiple Mixed Model Linear Regressions}

\begin{table}[ht!]
\centering
\footnotesize
\begin{tabular*}{\textwidth}{l @{\extracolsep{\fill}} rrrrr}
  \toprule
\textbf{Parameter} & \textbf{\textit{B}} & \textbf{SE} & \textbf{95\% CI} & \textbf{\textit{t}} & \textbf{\textit{p}} \\ 
  \midrule
(Intercept) & 5.08 & 0.30 & [4.49, 5.66] & 16.87 & \textless \space 0.001 \\ 
  Train Type [Dat\_Aug] & -0.02 & 0.18 & [-0.38, 0.34] & -0.11 & 0.916 \\ 
  Train Type [Pseu\_Lab] & 1.18 & 0.18 & [0.82, 1.54] & 6.43 & \textless \space 0.001 \\ 
  Dataset [CACAPO (en)] & -0.25 & 0.33 & [-0.89, 0.40] & -0.74 & 0.458 \\ 
  Dataset [CACAPO (nl)] & -0.33 & 0.33 & [-0.98, 0.32] & -0.99 & 0.323 \\ 
  Dataset [WebNLG] & -0.43 & 0.31 & [-1.05, 0.18] & -1.37 & 0.170 \\ 
  Train Type [Dat\_Aug] $\times$ Dataset [CACAPO (en)] & 0.01 & 0.20 & [-0.39, 0.40] & 0.04 & 0.965 \\ 
  Train Type [Pseu\_Lab] $\times$ Dataset [CACAPO (en)] & -0.34 & 0.20 & [-0.74, 0.06] & -1.69 & 0.092 \\
  Train Type [Dat\_Aug] $\times$ Dataset [CACAPO (nl)] & 0.00 & 0.20 & [-0.39, 0.40] & 0.02 & 0.982 \\ 
  Train Type [Pseu\_Lab] $\times$ Dataset [CACAPO (nl)] & 0.01 & 0.20 & [-0.39, 0.41] & 0.06 & 0.953 \\ 
  Train Type [Dat\_Aug] $\times$ Dataset [WebNLG] & 0.05 & 0.19 & [-0.33, 0.43] & 0.26 & 0.794 \\ 
  Train Type [Pseu\_Lab] $\times$ Dataset [WebNLG] & -0.08 & 0.19 & [-0.46, 0.29] & -0.44 & 0.661 \\ 
   \bottomrule
\end{tabular*}

\begin{tabular*}{\textwidth}{l @{\extracolsep{\fill}} rrrrr}
  \toprule
\textbf{Parameter} & \textbf{\textit{B}} & \textbf{SE} & \textbf{95\% CI} & \textbf{\textit{t}} & \textbf{\textit{p}} \\ 
  \midrule
(Intercept) & 4.25 & 0.32 & [3.63, 4.87] & 13.41 & \textless \space 0.001 \\ 
  Train Type [Dat\_Aug] & -0.05 & 0.19 & [-0.41, 0.32] & -0.26 & 0.795 \\ 
  Train Type [Pseu\_Lab] & -0.39 & 0.19 & [-0.75, -0.03] & -2.11 & 0.035 \\ 
  Dataset [CACAPO (en)] & 0.68 & 0.35 & [0.00, 1.36] & 1.95 & 0.051 \\ 
  Dataset [CACAPO (nl)] & 0.90 & 0.35 & [0.21, 1.59] & 2.57 & 0.010 \\ 
  Dataset [WebNLG] & -0.26 & 0.33 & [-0.90, 0.39] & -0.77 & 0.440 \\ 
  Train Type [Dat\_Aug] $\times$ Dataset [CACAPO (en)] & 0.04 & 0.20 & [-0.36, 0.44] & 0.22 & 0.830 \\ 
  Train Type [Pseu\_Lab] $\times$ Dataset [CACAPO (en)] & 1.04 & 0.20 & [0.64, 1.44] & 5.08 & \textless \space 0.001 \\ 
  Train Type [Dat\_Aug] $\times$ Dataset [CACAPO (nl)] & 0.05 & 0.20 & [-0.35, 0.45] & 0.24 & 0.806 \\ 
  Train Type [Pseu\_Lab] $\times$ Dataset [CACAPO (nl)] & 1.00 & 0.20 & [0.59, 1.40] & 4.86 & \textless \space 0.001 \\ 
  Train Type [Dat\_Aug] $\times$ Dataset [WebNLG] & 0.06 & 0.19 & [-0.32, 0.44] & 0.32 & 0.749 \\ 
  Train Type [Pseu\_Lab] $\times$ Dataset [WebNLG] & 2.03 & 0.19 & [1.65, 2.41] & 10.48 & \textless \space 0.001 \\
   \bottomrule
\end{tabular*}

\begin{tabular*}{\textwidth}{l @{\extracolsep{\fill}} rrrrr}
  \toprule
\textbf{Parameter} & \textbf{\textit{B}} & \textbf{SE} & \textbf{95\% CI} & \textbf{\textit{t}} & \textbf{\textit{p}} \\ 
  \midrule
(Intercept) & 3.06 & 0.14 & [2.79, 3.32] & 22.43 & \textless \space 0.001 \\ 
  Train Type [Dat\_Aug] & -0.07 & 0.09 & [-0.26, 0.11] & -0.77 & 0.439 \\ 
  Train Type [Pseu\_Lab] & 0.56 & 0.09 & [0.37, 0.74] & 5.95 & \textless \space 0.001 \\ 
  Dataset [CACAPO (en)] & -0.15 & 0.15 & [-0.45, 0.14] & -1.02 & 0.306 \\ 
  Dataset [CACAPO (nl)] & -0.22 & 0.15 & [-0.52, 0.07] & -1.47 & 0.141 \\ 
  Dataset [WebNLG] & -0.14 & 0.14 & [-0.41, 0.14] & -0.95 & 0.340 \\ 
  Train Type [Dat\_Aug] $\times$ Dataset [CACAPO (en)] & 0.10 & 0.10 & [-0.10, 0.30] & 0.96 & 0.335 \\ 
  Train Type [Pseu\_Lab] $\times$ Dataset [CACAPO (en)] & -0.16 & 0.10 & [-0.37, 0.04] & -1.59 & 0.111 \\ 
  Train Type [Dat\_Aug] $\times$ Dataset [CACAPO (nl)] & 0.02 & 0.10 & [-0.18, 0.23] & 0.23 & 0.820 \\ 
  Train Type [Pseu\_Lab] $\times$ Dataset [CACAPO (nl)] & 0.10 & 0.10 & [-0.10, 0.30] & 0.95 & 0.342 \\ 
  Train Type [Dat\_Aug] $\times$ Dataset [WebNLG] & 0.06 & 0.10 & [-0.13, 0.26] & 0.66 & 0.508 \\ 
  Train Type [Pseu\_Lab] $\times$ Dataset [WebNLG] & -0.04 & 0.10 & [-0.23, 0.15] & -0.45 & 0.656 \\ 
   \bottomrule
\end{tabular*}
\renewcommand{\thetable}{C.1}
\caption{Multiple Mixed Model Linear Regressions of Fluency, Correctness, and Grammaticality, respectively.}
\label{mmstats}
\end{table}

\end{document}